# A novel active learning framework for classification: using weighted rank aggregation to achieve multiple query criteria

Yu Zhao [a], Zhenhui Shi [a], Jingyang Zhang [a], Dong Chen [a], Lixu Gu [a], *

[a] *Department of Biomedical Engineering, Shanghai Jiao Tong University, Shanghai, China*

**Abstract:**

Multiple query criteria active learning methods have a higher potential performance than conventional active learning methods in which only one criterion is deployed for sample selection. A central issue related to multiple query criteria active learning methods concerns the development of an integration criteria strategy that makes full use of all criteria. The conventional integration criteria strategies adopted in relevant research all facilitate the desired effects, but several limitations still must be addressed. For instance, some of the strategies are not sufficiently scalable during the design process, and the number and type of criteria involved are dictated. Thus, it is challenging for the user to integrate other criteria into the original process unless modifications are made to the algorithm. Other strategies are too dependent on empirical parameters, which can only be acquired by experience or cross-validation and thus lack generality; additionally, these strategies are counter to the intention of active learning, as samples need to be labeled in the validation set before the active learning process can begin.

To address these limitations, we propose a novel multiple query criteria active learning method for classification tasks that employs a third strategy via weighted rank aggregation. The proposed method serves as a heuristic means to select high-value samples of high scalability and generality and is implemented through a three-step process: (1) the transformation of the sample selection to sample ranking and scoring, (2) the computation of the self-adaptive weights of each criterion, and (3) the weighted aggregation of each sample rank list. Ultimately, the sample at the top of the aggregated ranking list is the most comprehensively valuable and must be labeled. Several experiments generating 257 wins, 194 ties and 49 losses against other state-of-the-art multiple query criteria-based methods are conducted to verify that the proposed method can achieve superior results.

---

* Correspondence to: Image-Guided Surgery and Virtual Reality Lab, School of Biomedical Engineering, Shanghai Jiao Tong University, Dongchuang Road 800, Shanghai 200240, China
E-mail addresses: lereinion@163.com (Yu Zhao), shizhenhui90@gmail.com (Zhenhui Shi), J.Y.Zhang@sjtu.edu.cn (Jingyang Zhang), chendong8707@126.com (Dong Chen), gulixu@sjtu.edu.cn (Lixu Gu).

# 1 Introduction

*1.1. Motivation*

Active learning (AL) is a subfield of machine learning technology that is used to minimize the amount of annotation work that must be executed before training an accurate classification or regression model [1]. AL methods are unique in their use of various sample query criteria (SQC). These methods can help the user select a fraction of most 'valuable' samples for querying labels from massive volumes of unlabeled data [2-4]. On the basis of differences in the definition of 'valuable', AL methods can be broadly divided into two categories: representativeness and informativeness measure-based approaches [5-6].

As illustrated in Fig. 1, comparative studies [5,7] of AL methods have shown that most representativeness measure-based AL methods perform better when the number of labeled samples is few, whereas others, especially those that are informativeness measure-based, will usually overtake the former after substantial sampling. In this paper, the above phenomenon is referred to as 'the timeliness of AL'. The main explanation for this phenomenon is that representativeness measure-based AL methods can obtain the entire structure of a database upon their first use. However, these AL methods are not sensitive to samples that are close to the decision boundary, notwithstanding the fact that such samples are probably more important to the prediction model. In addition, informativeness measure-based AL methods always search for 'valuable' samples around the current decision boundary, and the optimal decision boundary cannot be found unless a certain number of samples have already been labeled [8]. In other words, the single query criterion can only guarantee its optimal performance over a period of time in the entire AL process, and the optimal period differs for each criterion.

Considering the above complementary characteristics, recent research reports have similarly proposed that the AL method could likely be improved if more than one SQC were deployed through one AL process to leverage the strengths of all methods [9]. The multiple query criteria AL method (MQCAL) has been developed for this purpose. The MQCAL method can combine most complementary information for each SQC through a special integration criteria strategy. A small number of samples that meet all involved criteria are selected for querying labels. The MQCAL method can in theory be more reliable, effective and resistant to interference because it takes several factors into account rather than focusing on one selection criterion, as is done in conventional AL methods. However, there are still some limitations that exist in MQCAL that require further effort to resolve (e.g., manual weight setting, the impossible combination), which will be described in detail below.

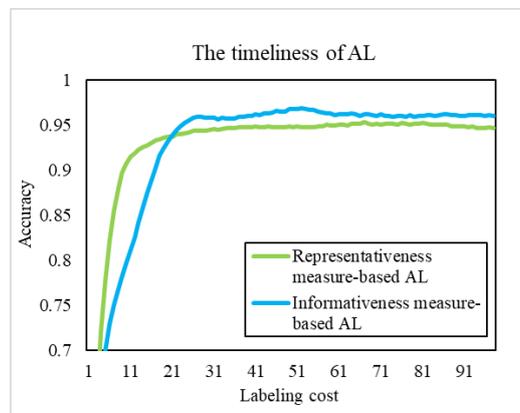

**Fig. 1.** The timeliness of AL

*1.2. Related work*

Most existing active learning methods are based on a single query criterion. Representativeness and informativeness measure-based AL methods are the two main branches of single criterion-based AL methods as shown in Fig. 2. The AL algorithms in the first category rely on the native data structure, and the samples that represent the majority of all samples are regarded as the most representative. According to the data structure expression, representativeness-based AL methods can be further subdivided into three classes that include Clustering Analysis (e.g., *Cluster* [6,10]), Sample Connection (e.g., *Diversity* [11], *Dissimilarity* [5], *Density* [12]), and Experimental Design (e.g., *TED* [13], *MAED* [14], *Random Walks* [15]). In contrast, informativeness measure-based AL methods always select a sample that has a high degree of uncertainty or is able to impart the greatest change to the current model. Based on the number of involved models, this method can also be further subdivided into two classes that include Certainty Based (e.g., *Margin* [1,16], *Entropy* [17], *EER* [18]) and Committee Based (e.g., *QBC* [19] and *Multiple View* [20]).

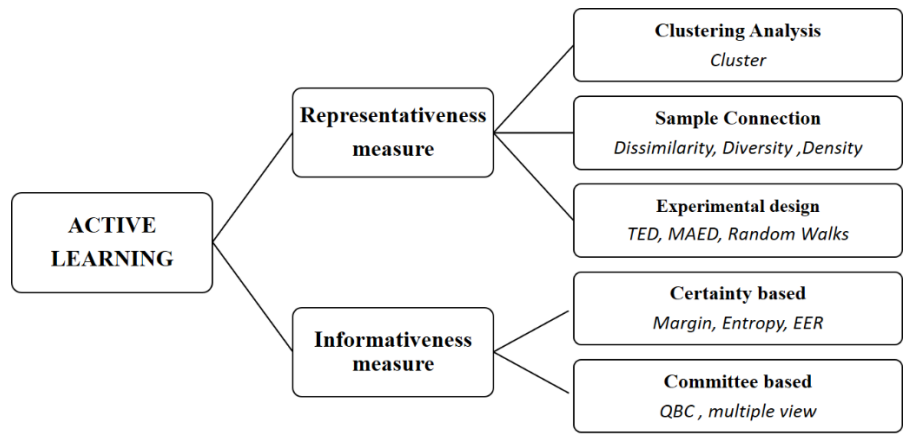

**Fig. 2.** The categories of traditional AL methods based on a single criterion

Compared with the traditional single criterion-based AL methods, existing research about the MQCAL is relatively sparse. Across these few studies, the selection and design of appropriate SQC for combining are usually their main foci of research rather than how to integrate all involved SQC together. After a careful review of existing methods, only four kinds of integration criteria strategies have been found as shown in Fig. 3.

Baram Y et al. [21] proposed the earliest form of the MQCAL method, as shown in Fig. 3 (A). For each iteration of this MQCAL, only one of the involved SQC with the highest criterion selection parameters is applied to choose samples. The criterion selection parameter is a variant of the multi-armed bandit algorithm proposed in [22]. Lughofer E [6] designed a two-phase AL process. In the first phase, the most representative samples based on clustering are selected, and a certainty-based AL approach is applied in the second phase. These MQCAL methods are beneficial primarily in terms of their high levels of efficiency. However, since only one involved query criterion is used in each iteration, their integration criteria strategies are more like criteria selection rather than criteria integration; hence, we refer to such strategies as 'CSAL' for short in this paper.

Shen D et al. [23] developed two other integration criteria strategies: parallel-form (shown in Fig. 3 (B)) and serial-form (shown in Fig. 3 (C)), both of which have been widely used in subsequent studies.

Serial-form MQCAL ('SMQCAL' for short) employs each SQC to select a certain number of samples from the selection results of the previous SQC in sequence as a multilayer filter. On the basis of Shen D's work [23], previous reports [4,11] further developed this approach by combining clustering and uncertainty-based SQC. Another report [24] applied this method to connect a K Nearest Neighbor-based cluster algorithm, an SVM margin algorithm and a genetic algorithm to propose an improved active learning method for hyperspectral image classification. Moreover, B. Demir et al. [25] proposed SMQCAL for remote sensing images, which involves SQCs based on uncertainty and diversity. Similar work includes a paper [26] in which samples in low-density regions were selected among the most uncertain samples; low-density regions are determined by exploiting the topological properties of SOM. This approach achieved fast convergence and performed well in both real multispectral remote sensing image classification tasks and hyperspectral remote sensing image classification tasks. In addition to the above classification tasks, B. Demir et al. [27] demonstrated that their serial-form MQCAL framework can perform well in regression tasks by efficiently identifying most of the diverse samples from high-density regions.

SMQCAL is efficient and operable and is widely used to address practical problems. In addition, the user can directly add several additional SQC on the basis of the original process, which can be regarded as strongly scalable. However, SMQCAL relies too heavily on two important settings, including the sequence of the applied SQC and the number of samples selected from each layer ($N_i$ in Fig. 3(C)), which are not generalized.

Parallel-form MQCAL ('PMQCAL' for short) can select optimal samples with regard to two different SQC using a weighted-sum optimization function. Based on this characteristic, previous studies [23,28] have effectively combined uncertainty and diversity to name entity recognition and natural language processing tasks, respectively. In addition, Huang H et al. [5] also employed the weighted-sum optimization function to combine the early stage-based SQC with the representativeness measure-based SQC (dissimilarity) for acquiring satisfactory AL selection results. Other similar studies include recent papers [29-30]. Although the respective criteria used to measure the values of the samples in each are not same, they all yield satisfactory results using the same basic mechanism, as shown in Fig. 3(B): the weight parameters $w_1$ and $w_2$ are used to balance the trade-off between each involved SQC. It is worth mentioning that, although the integration strategy in paper [31] is rendered as the product of two involved SQC with high exponent, it still can be regarded as the deformation of a weighted-sum optimization function and classified as a PMQCAL. To further improve parallel MQCAL, Huang S J et al. [8] developed another systematic way of measuring and combining representativeness and informativeness in the same SVM framework using the min-max AL view. This technique can be regarded as a state-of-the-art MQCAL with strong theoretical capacities.

However, PMQCAL also has two limitations. First, PMQCAL is not scalable. Thus, it is challenging for the user to integrate other SQCs into the original process unless modifications are made to the algorithm. Even so, the optimization function of this extended version may be unsolvable. Second, PMQCAL also places too much reliance on weight parameters. Using the wrong settings can result in suboptimal performance. Most of the above papers suggest that the user can directly use their recommended value [23,28] or obtain the optimal weights of a through cross-validation [29-30]. It is clear that it is also not generalizable for different applied data sets and is even slightly contrary to the original ideal of active learning, because the user should prepare some extra labeled samples as a validation set in the cross-validation process. In light of this problem, our group recently published an article [32] designing a double-strategy active learning method that is useful for mammographic mass classification, in which the combination weight is selected from predefined candidate values. Of course, this approach is not the best solution because it lacks a fine-tuning procedure. Similarly, due to the expectation maximization concept behind it, this solution is only designed for two SQC, further contributing to a lack of scalability.

Additionally, Wang et al. [9] proposed a fourth MQCAL, as shown in Fig. 3 (D), by transforming the problem of integration criteria into a multicriteria decision-making system (termed 'MCDMAL' here), which also yields good results in the multiple-instance learning environment rather than in a classification task. However, this method has high algorithm complexity. Its implementation and execution are quite difficult, and not every kind of SQC can be integrated into their MQCAL process.

Furthermore, an unsolved problem concerns the establishment of a mechanism that allows for a dynamic and adaptive tradeoff between each SQC that is used for each AL iteration [8,11]. This problem is addressed in most of the above works as a suggested avenue for future research. To our knowledge, only Donmez P et.al. [7] has proposed a means of tuning the weights of two SQC in each iteration by calculating the estimated future residual error reduction level.

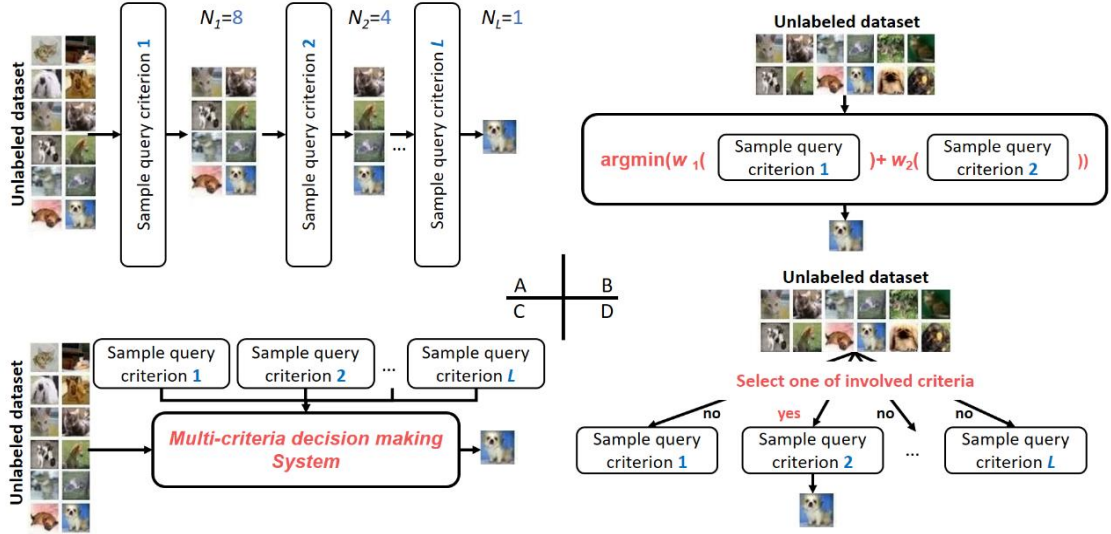

**Fig. 3.** The existing MQCAL process (A: the process of CSAL, B: the process of PMQCAL, C: the process of SMQCAL, D: the process of MCDMAL)

*1.3. Our approach - the main concept*

We realized that the sample selection problem in AL methods is also a sample ranking problem and were inspired by the recommendation technologies that have developed in recent years. Hence, in this manuscript, we develop a novel weighted rank aggregation based MQCAL for classification tasks, which can be regarded as a fifth form of the integration criteria strategy, which we term 'RMQCAL'.

To implement the proposed method, three additional steps are added to the framework of the original AL process, as shown in Fig. 4. In any iteration of the AL process, all involved SQC first need to be tweaked in order to invert the problem of sample selection into sample ranking and scoring. Next, every pair of ranking and scoring lists based on their corresponding SQC can be obtained from the remaining unlabeled samples (see Section 2.2). Third, using the best-versus-second-best (BVSB) strategy, the weights of each SQC for every iteration of the AL process can be dynamically obtained from the current score lists (see Section 2.3). Then, the rank lists of all SQC involved are weighted and combined as a comprehensive ranked list through our improved weighted rank aggregation method (see Section 2.4). The sample ranked highest in this comprehensive ranked list is then considered to be the most comprehensively valuable and the most in need of labeling for this iteration.

The innovation of this article manifests in both the originality of the study object and the proposed solution. The study object of the proposed RMQCAL focuses on the design of an integration criteria strategy that can integrate each SQC involved rather than designing several specific SQCs and adding them together using empirical weight-parameter settings. Moreover, the solution of the proposed

RMQCAL treats criteria integration as a special rank aggregation problem to be solved using a Markov chain; this methodology differs completely from that of earlier studies. In terms of the algorithm itself, RMQCAL has the following advantages over the existing MQCAL. **Scalability:** Similar to serial-form MQCAL, any number and type of SQC can be easily introduced into our RMQCAL process without establishing a more complex optimization function. **Uniformity:** The uniformity of each SQC can be guaranteed by converting sample selection, the purpose of each SQC, into sample ranking. **Generality:** Our method no longer employs any empirical parameters; instead, each tradeoff behind SQC is self-adaptive. **Dynamics:** As in most papers, except in their future work, the tradeoffs between each SQC used in our method are dynamic, and they change according to their differential contributions in each iteration of the AL process.

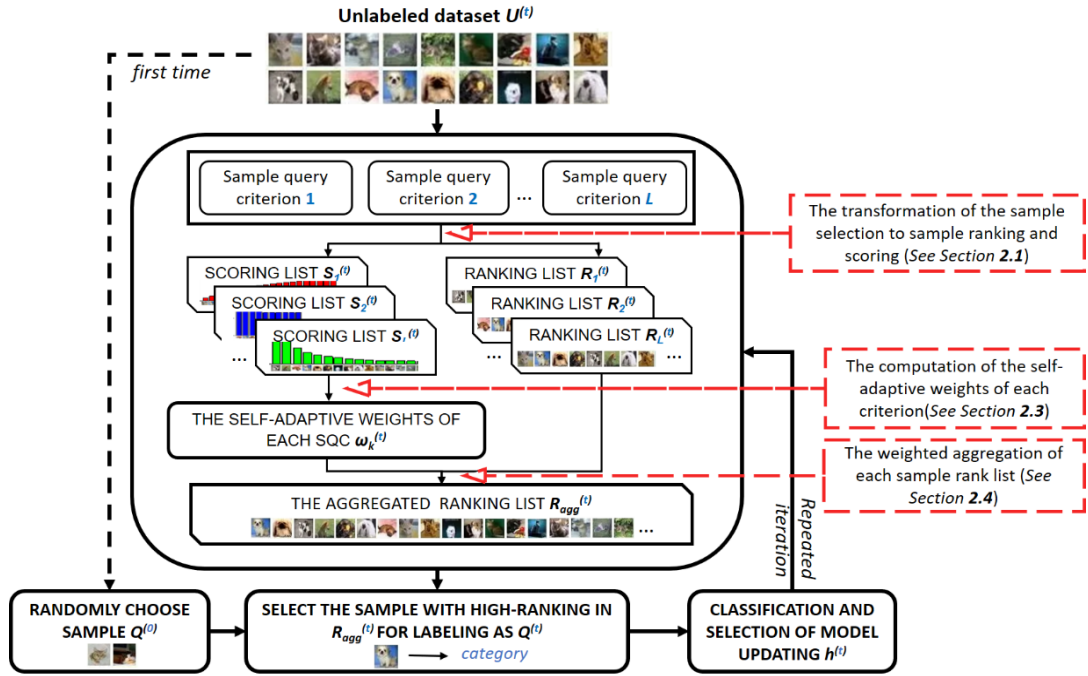

**Fig. 4.** The process of the proposed MQCAL based on rank aggregation (red boxes indicate the major steps of the proposed method)

Moreover, RMQCAL offers a potential predominance in practical applications. The aim of AL methods is to reduce the annotation work of unlabeled samples in hand. However, when dealing with an unlabeled dataset in real-word problems, in order to select the most appropriate AL method and acquire optimal empirical parameters, it remains unavoidable for a certain number of labeled samples to be required to establish a validation set. For the proposed method, because it requires no empirical parameters and has high scalability, the users merely have to employ all candidate AL methods for selection as the multiple criteria in the proposed method. RMQCAL can satisfactorily combine them as an optimal ensemble AL method with self-adaptive adjustment of weights. The validation set is no longer needed, which can better serve the needs presented by practical problems, especially when the labeling cost of each sample is very expensive.

The highlights of this work include the following. (1) To the best of our knowledge, this is the first work to analyze and induct the existing MQCAL method with different integration criteria strategies. (2) This is also the first work to implement the MQCAL method by introducing weighted rank aggregation approaches, and the proposed framework may inspire future AL. (3) We present a mechanism that allows for a dynamic and self-adaptive tradeoff between any number and kind of involved SQC in a unified system by introducing the BVSB strategy. (4) We summarize basic rules for the use of our RMQCAL. The potentially best combination of involved SQC and rank aggregation approaches is also found from experimental comparative results. (5) Several comparative experiments are conducted to prove the effectiveness of the proposed RMQCAL method in many public data sets.

The remainder of this paper is organized as follows. In Section 2, the framework of our RMQCAL is presented, and the three main steps of this framework are discussed in detail. Section 3 describes the experiments that were conducted to evaluate the performance of our RMQCAL and to define optimal operation parameters. In turn, the optimum combination of SQC and best methods for rank aggregation can be obtained. Finally, our conclusions are presented in Section 4.

## 2 Approach
### 2.1. Problem definition

Assume that there is an initial dataset $D$ that is used to train a binary classification model with a lower labeling cost. In any iteration of AL process $t$, the entire dataset $D$ is always divided into two subsets: the subset $A^{(t-1)}$ and $U^{(t)}$. $U^{(t)}$ is the currently unlabeled data-set, which stores $|U^{(t)}|$ unlabeled sample $u_n^{(t)}$ in the form of feature vector, where $n \in [1, \cdots, |U^{(t)}|]$, and $|.|$ is a function that is used to calculate the length of an array. In addition, the existing labeled dataset is defined as $A^{(t-1)}$ and is obtained from the previous iteration, which also stores the feature vector of labeled samples as $[a_m^{(t)}, y_m]$, where $m \in [1, \cdots, |A^{(t-1)}|]$ and $y_m = \{1, -1\}$. Through one specialized SQC $F^{(t)}(.)$ from the old learning model $h^{(t-1)}$ that was previously trained, a conventional single criterion-based AL method selects several of the most important samples $Q^{(t)}$ with the highest value from $U^{(t)}$ in each iteration. Then, the labeled dataset can be reconstituted and used to train a new $h^{(t)}$ and update the SQC as $F^{(t+1)}(.)$ for the next iteration $t+1$ ($A^{(t)} = A^{(t-1)} \cup Q^{(t)}$).

Most of the SQC $F^{(t)}(.)$ in the conventional AL process can be described as in formula (1):

$$Q_{select}^{(t)} = F^{(t)}(U^{(t)}, N) = \underset{V \subset U^{(t)}}{argmin} \sum_{n=1}^{N} f^{(t)}(v_n) \tag{1}$$

where $u_n^{(t)}$ are the elements in $V$ ($v_n \in V$); $f^{(t)}(.)$, which is the kernel function in this SQC $F^{(t)}(.)$ that is used to calculate the score of every unlabeled sample for sample selection, according to the existing labeled samples $A^{(t-1)}$; $N$ is the number of selections in each iteration of the AL process; $|V| = N$, which is usually set as 1.

Unlike the traditional AL process, the intermediate process of MQCAL involves the use of a combination of $L$ SQC, namely, $F_k^{(t)}(.)$ $n \in [1, \cdots, L]$. Different $F_k^{(t)}(.)$ have different kernel function $f_k^{(t)}(.)$. Only the most comprehensively valuable samples that meet all these SQC are selected for labeling in each loop iteration. The kernel of MQCAL is used to establish an integration criteria strategy that can combine most complementary information of each SQC as $\bigwedge_{k=1}^{L}(.)$. With regard to the existing MQCAL, including those that are criteria selection-based, MCDM system-based, parallel-form and serial-form, each of their integration criteria strategy can be calculated as formula (2), formula (3), formula (4), and formula (5), respectively:

$$Q_{cs}^{(t)} = \bigwedge_{k=1}^{L}(F_k^{(t)}(U^{(t)}, N)) = \underset{V \subset U^{(t)}}{argmin} \sum_{n=1}^{N} f_{k^*}^{(t)}(v_n) \tag{2}$$

where $k^* = \underset{k \in [1,\ldots,L]}{argmax}(CSP_k^{(t)})$, $CSP_k^{(t)}$ is the criteria selection parameter of $F_k^{(t)}(.)$ in $t$ th iteration.

$$Q_{MCDM}^{(t)} = \bigwedge_{k=1}^{L}(F_k^{(t)}(U^{(t)}, N)) = \underset{V \subset U^{(t)}}{argmin} \sum_{n=1}^{N} info^{(t)}(v_n) \tag{3}$$

where $info^{(t)}$ represents the difference between dominated index and the dominating index of each sample calculated by MCDM system and $F_k^{(t)}(.)$; where $w_k$ is the weight parameter of $F_k^{(t)}$. Weight parameters are always fixed empirically or through cross-validation.

$$Q_{parallel}^{(t)} = \wedge_{k=1}^{L}(F_k^{(t)}(U^{(t)}, N)) = \underset{V \subset U^{(t)}}{argmin} \sum_{n=1}^{N} \sum_{k=1}^{L} w_k f_k^{(t)}(v_n) \quad (4)$$

$$Q_{serial}^{(t)} = \wedge_{k=1}^{L}(F_k^{(t)}(U^{(t)}, N)) = F_L^{(t)}\left(\left(\cdots\left(F_2^{(t)}\left(F_1^{(t)}(U^{(t)}, N_1)\right), N_2\right)\cdots\right), N_L\right) \quad (5)$$

where $N_k$ is the number of selections in layer $k$, and $N_L = N$.

Both their advantages and disadvantages are mentioned in the previous section.

We noted that for each SQC $F_k^{(t)}(.)$, the corresponding scoring list $S_k^{(t)}$ of the currently unlabeled dataset $U^{(t)}$ can be calculated by the corresponding kernel function $f_k^{(t)}(.)$, as given by formula (6):

$$S_k^{(t)} = \left[f_k^{(t)}(u_1^{(t)}), \dots, f_k^{(t)}(u_n^{(t)}), \dots, f_k^{(t)}\left(u_{|U^{(t)}|}^{(t)}\right)\right] \quad (6)$$

Meanwhile, the ranking list $R_k^{(t)}$ of each sample in $U^{(t)}$ can also be easily obtained by sorting $S_k^{(t)}$ in ascending or descending order. Then, we suggest that the integration criteria strategy of our RMQCAL can be designed as formula (7),

$$Q_{RMQCAL}^{(t)} == F^{(t)}(U^{(t)}, N) = \underset{V \subset U^{(t)}}{argmin} \sum_{n=1}^{N} R_{agg}^{(t)}(v_n^{(t)}) \quad (7)$$

where $R_{agg}^{(t)}$ is the aggregated ranking list that satisfies formula (8), K is the calculation of Kendall's tau or Spearman's footrule distance [33], and $w_k$ is the self-adaptive tradeoff of $R_k^{(t)}$, which is calculated by $S_k^{(t)}0$

$$R_{agg}^{(t)} = \underset{R}{argmin} \frac{1}{L} \sum_{k=1}^{L} w_k K(R, R_k^{(t)}) \quad (8)$$

Then, the problem of our RMQCAL can be transformed as a weighted rank aggregation problem. In other words, three core contents of RMQCAL include the acquisition of $S_k^{(t)}$ and $R_k^{(t)}$, the weighted computation of each criterion $\omega_k$ and how to effectively combine each SQC together. These will be individually discussed in the following three parts.

*2.2. The transformation of the sample selection to sample ranking and scoring*

The purpose of this step is to obtain $L$ different pairs of rank and score lists $S_k^{(t)}$ and $R_k^{(t)}$ of all remaining unlabeled samples in $U^{(t)}$ from the $L$ different SQC. Because all SQC can be described as shown in formula (1), the scoring lists of currently unlabeled dataset $S_k^{(t)}$ can be further denoted as formula (6), and the ranking lists $R_k^{(t)}$ are the ranking of their corresponding $S_k^{(t)}$ from small to large.

Taking the SQC of some typical AL methods as examples, given that they are also important components of our RMQCAL and the control methods used in this paper, their separate scoring kernel functions $f_k^{(t)}(.)$ are written as the following: formula (9), formula (10), formula (11) and formula (12).

**Margin-based SQC:** [16]

$$f_{margin}^{(t)}(x) = P(y_{max}^* | x, h^{(t-1)}) \quad (9)$$

**Diversity-based SQC:** [11]

$$f_{diversity}^{(t)}(x) = \max\left(\cos^{-1}\left(\frac{K(x, a_m^{(t-1)})}{\sqrt{K(x,x)K(a_m^{(t-1)}, a_m^{(t-1)})}}\right)\right) \quad (10)$$

**QBC-based SQC:** [19]

$$f_{QBC}^{(t)}(x) = -1 \times \sigma(h_1^{(t-1)}(x), \dots, h_g^{(t-1)}(x)) \tag{11}$$

**TED-based SQC:** [13]

$$f_{TED}^{(t)}(x) = -1 \times \sum_{i=1}^{|U^{(t)}|} Z(:,i), \quad i \text{ is the position of } x \text{ in } U^{(t)} \tag{12}$$

where $y_{max}$ is the most likely label of x, κ is the kernel distance, $g$ is the number of committees in QBC,

$$Z = \min_{Z} \lVert U^{(t)} - U^{(t)}Z \rVert_{2,1} + \lambda \lVert Z^T \rVert_{2,1}, \ s.t. Z = [z_1, \dots, z_{|U^{(t)}|}] \in R^{|U^{(t)}| \times |U^{(t)}|} \text{ and } \sigma \text{ is the function}$$

used to calculate the standard deviation.

Because each SQC has a different AL concept, each $S_k^{(t)}$ must be normalized from -1.0 to 1.0 and sorted from smallest to largest as $S_k^{*(t)}$. The following three points are worth mentioning. (1): In our RMQCAL, we force the definition that the score of the sample with the highest value is the lowest in the scoring list. An SQC that does not conform to the above definition should be revised by multiplying it by -1. (2) When experimental design-based criteria are included in the involved SQC, their corresponding rank and score lists need to be calculated only once before the first iteration as $R_k^{(0)}$ and $S_k^{*(0)}$ because the experimental design-based SQC does not involve model updating, and its score list is constantly in subsequent iterations. (3) $R_k^{(t)}$ from committee-based SQC includes numerous duplicate values; thus, tie conditions applicable for obtaining $R_k^{(t)}$ should be reflected rather than assigned random rankings.

*2.3. The computation of the self-adaptive weights of each criterion*

Based on the timeliness of the AL noted above, the contributions of each criterion change in response to different stages of the AL process. The subjective definition of each kind of SQC as its weight, which follows from the old pattern of the serial-designed MQCAL, is not used. In our method, a dynamic weighting system is established for calculating the self-adaptive weights of each SQC for every iteration. Inspired by image retrieval [34], weights can be calculated from the distribution of score lists $S_k^{*(t)}$. Intuitively, we expect that the $S_k^{*(t)}$ of the best SQC should satisfy the formula (13) and that the best SQC will look like the red bar shown in Fig. 5(a) if we here assume that the number of sample selections equals $N = 3$.

$$S_{best}^{*(t)}(index) = \begin{cases} \min(S_{best}^{*(t)}), & if \ index \leq N \\ \max(S_{best}^{*(t)}), & otherwise \end{cases}, index = 1, 2, \dots, |U^{(t)}| \tag{13}$$

where $S_k^{*(t)}(index)$ is the score of the sample corresponding to *index* in $S_k^{*(t)}$.

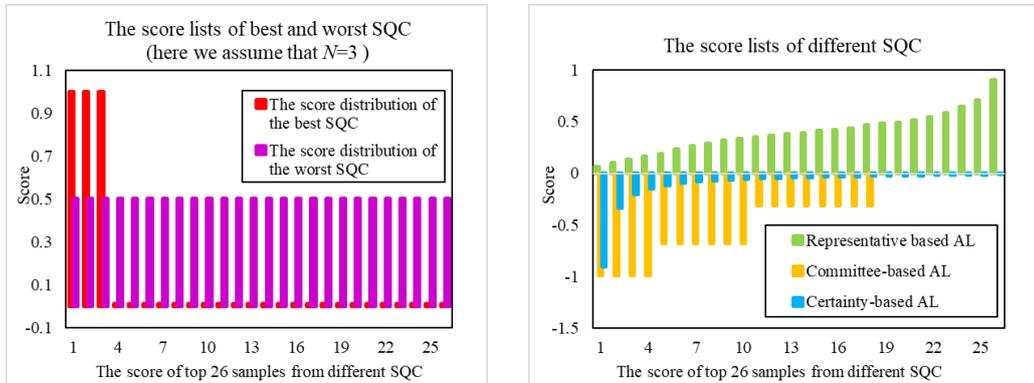

**Fig. 5.** Score lists of several SQCs (intercepting only the scores of the top 26 most valuable samples defined by three AL methods). The figure on the left is the score list of the best and worst SQCs, and the figure on the right is the score list of the real SQCs.

For the opposite case, the worst $S_k^{*(t)}$ is a straight line represented by the purple bar shown in Fig. 5(a), wherein all values of $S_k^{*(t)}$ are completely equal. For practical issues, the real curves of $S_k^{*(t)}$ are typically displayed as shown in Fig. 5(b), and the $S_k^{*(t)}$ values of certainty and representativeness measure-based AL are curvilinear and contain fewer duplicate score values. In contrast, the $S_k^{*(t)}$ of a committee-based AL is ladder-like and has several duplicate score values.

Exploiting a similar trick for the BVSB algorithm and the three curves of the different score lists described above, we suggest that the more obvious the difference in one of the $S_k^{*(t)}$ between the selected samples and the almost-selected samples, the greater the contribution of its corresponding SQC. For this, we present a weight-assignment method, as illustrated as formula (14), if none of the involved SQC is committee-based.

$$w1_k^{(t)} = (S_k^{*(t)}(N) - S_k^{*(t)}(N+1))/(S_k^{*(t)}(1) - S_k^{*(t)}(|U^{(t)}|)), \ k \in \text{not committee} \tag{14}$$

For the score list of committee-based SQC, in which the shapes of score lists are entirely different from those in the two other cases, the above formula (14) cannot be applied here because its unique distribution ($S_k^{*(t)}(N)$ is likely to be equal to $S_k^{*(t)}(N+1)$). Then, we believe that the weights between each committee-based SQC can be calculated using formula (15), as shown below, instead. This means that the SQC, which has lower likelihoods of the selected and almost-selected samples sharing the same score, should be assigned a higher weight. I(.) is an indicator function that is equal to one if conditions within the parentheses are satisfied; otherwise, it is equal to zero.

$$w2_k^{(t)} = \sum_{index=N+1}^{|U^{(t)}|} \frac{I(S_k^{*(t)}(index) \neq S_k^{*(t)}(N))}{|U^{(t)}|}, \ k \in \text{committee} \tag{15}$$

However, we have not developed a more generalized weight-assignment method that applies to representativeness-, certainty- and committee-based SQC. When the involved SQC in MQCAL includes all three of the types listed above, the only workable revised weight assignment scheme is written as formula (16): where $c2$ is the number of committee-based SQC, and $c1 = L - c2$.

$$w_k^{(t)} = \begin{cases} \frac{c1}{L} * w1_k^{(t)} / \sum_j w1_j^{(t)}, & k, j \in \text{not committee} \\ \frac{c2}{L} * w2_k^{(t)} / \sum_j w2_j^{(t)}, & k, j \in \text{committee} \end{cases} \tag{16}$$

---

**Algorithm 1: Weight calculations of the RMQCAL process**

**Input:** The $L$ score lists $S_k^{*(t)}$ from the $n_1$ certainty-based SQC, $n_2$ committee-based SQC and $n_3$ representativeness-measure SQC, where $k = 1, 2, \cdots, L (L = n_1 + n_2 + n_3)$, and the number of samples is selected from each iteration $N$.

1: Normalize each score list to -1.0 to 0 or 0 to 1.0; then, $S_k^{*(t)}$ can be obtained by sorting the scores in ascending order.

2: Calculate two correction parameters: $c2 = n_2$, and $c1 = n_1 + n_3$.

3: Calculate $w_k^{(t)}$ from formulas (14), (15) and (16).

**Output:** A vector $\boldsymbol{w}^{(t)} = [w_1^{(t)}, \cdots, w_L^{(t)}]$ that represents the weight of each $L$ SQC in the $t$ th iteration.

---

*2.4: The weighted aggregation of each sample rank list*

After obtaining $R_k^{(t)}$ from step 1 and $\boldsymbol{w}^{(t)}$ from step 2, the following problem is similar to a rank aggregation problem that can be elegantly solved by using improved rank aggregation methods.

Here, it is useful to review rank aggregation methods. Lin S summarized existing rank aggregation methods that had been developed up until 2010 [33] and are used to address problems related to the recommendation system. In recent years, many methods of rank aggregation have been designed, including the following: Borda's method, Bucklin voting [35], the Markov chain [36], Thurstone's model, the cross-entropy Monte Carlo model [33], the Condorcet method [37] and other stochastic methods [38]. Rank aggregation is now widely employed to address information retrieval problems. To our knowledge, this is the first study to apply rank aggregation methods to the AL problem.

However, some differences remain between common rank aggregation problems and our MQCAL problem, and some of the rank aggregation methods may not properly address MQCAL problems. Therefore, before introducing these methods into our algorithm, they must still be selected and improved. Put simply, specific differences include the following: (1) the number of rank lists ($L$) is not sufficiently large to establish a statistical model; (2) the number of elements in $R_k^{(t)}$ is large, particularly for the first iterations (when $t$ is small), which can result in inefficiencies; (3) traditional rank aggregation problems seldom involve weighting; and (4) in most cases, $R_k^{(t)}$ is not a shuffled list from one to $|U^{(t)}|$ as the same rankings may be involved.

Point (1) implies that statistical model-based rank aggregation methods, e.g., Thurstone's model, cannot work. With regard to point (2), we apply rank aggregation methods for lower computing complexity, e.g., Borda's and Bucklin's methods and the Markov chain. In addition, a sample truncation method is proposed as a mean to further reduce the number of samples involved in rank aggregation. Regarding points (3) and (4), some improvements are made to existing rank aggregation methods (e.g., adding weights to each list).

For the above problems, we present three feasible means of rank aggregation of varying computing complexity and performance that are based on enhanced versions of the Borda, Bucklin voting and Markov chain approaches.

*2.4.1 Borda's methods*

Borda's methods are the most popular and intuitive rank aggregation methods [33], and they are still widely used to study elections. There are two main phases of Borda methods.

1. The first phase involves the construction of a mapping function $\text{MAP}(.)$ between the ranking $R_k^{(t)}$ and its corresponding Borda score $B_k^{(t)}$. When addressing practical issues, $\text{MAP}(.)$ is typically designed to score as 1 when ranked first, as 2 when ranked second, and so on. In the other words, $\text{MAP}(.)$ is expressed as the formula (17):

$$B_k^{(t)}(index) = MAP\left(R_k^{(t)}(index)\right) \approx R_k^{(t)}(index), index = 1,2,\ldots,|U^{(t)}| \qquad (17)$$

where $R_k^{(t)}(index)$ is the score of indexed samples in $R_k^{(t)}$.

For the application of such methods to our RMQCAL, due to the processing that is involved in weighting, formula (17) should be reformulated as formulas (18):

$$B_k^{(t)}(index) = MAP\left(R_k^{(t)}(index)\right) \cdot w_k^{(t)} \approx R_k^{(t)}(index) \cdot w_k^{(t)}, index = 1,2,\ldots,|U^{(t)}| \qquad (18)$$

2. The second phase involves the use of $f_{borda}(.)$, which we refer to as the Borda score fusion algorithm. This algorithm is used to obtain the overall Borda score $B_{borda}^{(t)}$ by combining all Borda scores in formula (19):

$$B_{borda}^{(t)} = f_{borda}\left(B_1^{(t)}, B_2^{(t)}, \ldots, B_L^{(t)}\right) = \begin{cases} min\left(B_1^{(t)}, B_2^{(t)}, \ldots, B_L^{(t)}\right) & (minimum) \\ median\left(B_1^{(t)}, B_2^{(t)}, \ldots, B_L^{(t)}\right) & (median) \\ (\prod_{k=1}^{L} B_k^{(t)})^{1/L} & (geometric\ mean) \\ \sum_{k=1}^{L} (B_k^{(t)})^p & (p-norm) \end{cases} \qquad (19)$$

where $p$ is typically set equal to 1 (Here, the p-norm algorithm is the arithmetic mean).

The Borda-based rank aggregation method of our RMQCAL involves the following four steps:

**Algorithm 2: Borda-based rank aggregation of the RMQCAL process**

Input: The L rank lists $R_k^{(t)}$ from corresponding SQC values, where k=1, 2, ..., L and the number of samples is selected in each iteration $N$ and $w^{(t)}$.

1: Determine the mapping function $MAP(.)$, the core fusion algorithm $f_{borda}(.)$, and parameter $N$

2: An $L \times |U^{(t)}|$ Borda score list can be established according to the above formula (18), in which every row is the Borda score of one rank list, and every column includes the $L$ Borda score of one sample from different SQC values.

3: The overall Borda score can be obtained from formula (19). For each column of the above list, $R_{agg}^{(t)}$ is the rank list of this Borda score from small to large.

4: $N$ samples with the lowest overall Borda score $B_{borda}^{(t)}$ can be selected as $Q^{(t)}$.

Output: $Q^{(t)}$, denoting the most $N$ valuable samples, is selected from $U^{(t)}$ in the *t*-th iteration.

*2.4.2. Bucklin voting method*

According to the method described in [35], the earliest iteration of the Bucklin voting method was also used in voting systems for candidate selection. According to the kernel principle of the Bucklin voting method, when one candidate has a majority, that candidate wins. Otherwise, the second choice is added to the first choice. Whether one candidate has a majority is re-estimated; if so, that candidate wins. If not, the previous tasks are repeated.

Due to the importance of weighting factors, the algorithm is similar to the Electoral College system. Election candidates are used as samples. The first and second choices correspond, respectively, to the first and second ranked $R_k^{(t)}$ values. $L$ SQC are no longer $L$ voters but are $L$ states, and $R_k^{(t)}$ can be regarded as the number of electoral votes cast in each state. Based on a previous publication [35], the overall process is described as follows.

**Algorithm 3: Bucklin voting-based rank aggregation of the RMQCAL process**

**Input:** $L$ rank lists $|U^{(t)}|$ from corresponding SQC, where k=1, 2, ..., $L$ and the number of samples is selected in each iteration $N$ and $w^{(t)}$.

1: Establish a $1 \times |U^{(t)}|$ sparse list $SL$, where each column records the electoral votes of each sample from $L$ rank lists initialized to zeros.

2: Set *ch*=1 and *ii*=1 and construct an empty $Q^{(t)}$.

   3: Start searching the sample with the *ii* th value in the aggregate rank list; the positioning of this sample is saved in $Q^{(t)}$.

**While** $(ii < 1 + N)$

   4: **For** *j*=1: $L$

      If the *index* th sample is the *ch* th of $R_k^{(t)}$, $R_k^{(t)}(index) = ch$

         Update the list $SL$, $SL(ch, index) = SL(ch, index) + w_k^{(t)}$.

      **End**

   **End**

   **If** there is a column *index\** of list $SL$ whose summation is greater than 0.5

      5: Record *index\** in $Q^{(t)}$, $ii = ii + 1$; clear this column to zero to prevent it from being selected a second time.

   **Else**

      5: $ch = ch + 1$; insert a new line below $SL$.

   **End**

**End**

**Output:** $Q^{(t)}$, which are the $N$ samples with the highest value selected from $Q^{(t)}$, of the *t* th iteration.

Theoretically, the Bucklin voting method is an ideal method to use in our RMQCAL because it makes no attempt to initially aggregate a complete rank list $\boldsymbol{R}_{agg}^{(t)}$ for all involved samples. The samples of each place in the rank list after aggregation are confirmed individually and are exactly what our RMQCAL requires (Only the most $N$ valuable samples need be selected for each iteration, where $N$ is always small).

*2.4.3. Markov chain method*

The Markov chain method was first introduced into the PageRank algorithm (a practical rank aggregation problem) by Dwork in 2001. As noted in the literature [36], the Markov chain serves as an elegant, rational and high-performance solution to rank aggregation problems. The kernel ideas of the conventional Markov chain that are used for rank aggregation typically involve two steps. Step 1: Convert aggregated targets by incorporating several input ranking lists into a specific transition matrix using one form of probability assignment $P(.)$. Step 2: According to [36], regardless of the initial state, the Markov chain system based on one specific transition matrix will always eventually reach a unique fixed point at which the state distribution does not change. We define this point as the stationary distribution of the corresponding transition matrix, which is also the basis for ranking lists after aggregation. According to the different transition matrices, the Markov chain method can be subdivided into the following: MC1, MC2 and MC3 [33].

However, traditional Markov chain methods are not completely suited to address our RMQCAL problem because the original methods do not apply weights. In addition, its computation complexity is high, particularly when $t$ is small. To solve these two problems, two changes are made to the Markov chain method in the proposed RMQCAL approach, as follows.

First, before building a transition matrix, an extra 'sample truncation' step is added to significantly reduce computation complexity levels using only some of the samples. Only the sample that are among the top $N^*$ in each $\boldsymbol{R}_k^{(t)}$ at least once will join the next phase of transition matrix establishment, and the remainder are ignored. The sample truncation process is expressed as formula (20):

$$\boldsymbol{R}_k^{*(t)} = \boldsymbol{R}_k^{(t)}(n), n \in \{i | \sum_{k=1}^{L} I(\boldsymbol{R}_k^{(t)}(i) \leq N^*) \geq 1, i = 1, \dots, |\boldsymbol{U}^{(t)}|\} \tag{20}$$

where $\boldsymbol{R}_k^{*(t)}$ is a modified version of $\boldsymbol{R}_k^{(t)}$ after sample truncation, $N^*$ is typically set as $N^* = N + tun_2$. ($tun_2$ can be set equal to 5), and $N < N^* \leq |\boldsymbol{R}_k^{*(t)}| \ll |\boldsymbol{R}_k^{(t)}| = |\boldsymbol{U}^{(t)}|$.

This improvement is applicable to our RMQCAL problem. For our RMQCAL, in each iteration, the samples that are used are ranked at the top of the $N$ list after rank aggregation. Furthermore, the higher a sample ranks in any $\boldsymbol{R}_k^{(t)}$, the more likely it is to occupy the top $N$ place in the aggregated list of all $\boldsymbol{R}_k^{(t)}$ values. Therefore, the rank aggregated results of each $\boldsymbol{R}_k^{(t)}$ and the front section of each $\boldsymbol{R}_k^{(t)}$ are likely to be the same, particularly when $N$ is not large and when most $\boldsymbol{R}_k^{(t)}$ values are relatively similar.

Second, for each pair of samples $u_i^{(t)}$, $u_j^{(t)}$ ($i \neq j$ and $i, j \in [1, 2, \cdots, |\boldsymbol{R}_k^{*(t)}|]$), the improved weighted transition probability $\boldsymbol{Tran}^{(t)}(i,j)$ in RMQCAL can be described as formula (21):

$$\boldsymbol{Tran}^{(t)}(i,j) = P(u_i^{(t)} \to u_j^{(t)}) = \begin{cases} \frac{1}{|U^{(t)}|} \cdot I\left(\sum_{k=1}^{L} w_k^{(t)} \cdot \left(I(\boldsymbol{R}_k^{(t)}(i) > \boldsymbol{R}_k^{(t)}(j))\right) > 0\right) : MC1 \\ \frac{1}{|U^{(t)}|} \cdot I\left(\sum_{k=1}^{L} w_k^{(t)} \cdot (I(\boldsymbol{R}_k^{(t)}(i) > \boldsymbol{R}_k^{(t)}(j))) > \frac{1}{2}\right) : MC2 \\ \frac{1}{|U^{(t)}|} \cdot \sum_{k=1}^{L} w_k^{(t)} \cdot \left(I(\boldsymbol{R}_k^{(t)}(i) > \boldsymbol{R}_k^{(t)}(j))\right) \quad\quad : MC3 \end{cases} \tag{21}$$

After all $P(u_i^{(t)} \to u_j^{(t)})$ values have been calculated, $P(u_i^{(t)} \to u_i^{(t)})$ can be obtained from formula (22):

$$\boldsymbol{Tran}^{(t)}(i,i) = (u_i^{(t)} \to u_i^{(t)}) = 1 - \sum_{i \neq j} \boldsymbol{Tran}^{(t)}(i,j) \tag{22}$$

Because $L$ is typically not large in our RMQCAL problem, the above $\textbf{Tran}^{(t)}$ is often a large sparse matrix with several 0 elements. To ensure ergodic results for the transition matrix, a tuning parameter $t$ is introduced and treated as follows in formula (23):

$$\textbf{Tran}^{*(t)}(i,j) = \textbf{Tran}^{(t)}(i,j) \times (1 - tun_1) + \frac{tun_1}{|U^{(t)}|} \tag{23}$$

where $tun_1$ is typically set to range from 0.01 to 0.15, as specified by the above reference.

Finally, from the perspective of matrix theory, the stationary distribution of one transition matrix is its principal left eigenvector, which can be computed from a regular power-iteration algorithm after transposing the above matrix. The improved MC method used for RMQCAL problem is as follows:

**Algorithm 4: Markov chain-based rank aggregation for the RMQCAL process**

**Input:** $L$ rank lists $\textbf{R}_k^{(t)}$ and corresponding $\textbf{w}^{(t)}$ values from the corresponding SQC, where $k$=1, 2, ..., $L$, the number of samples selected in each iteration is designated $N$, the tuning parameter is designated $tun_1$, and the number of other samples of interest is designated $tun_2$.

1: For each $\textbf{R}_k^{(t)}$ except for the committee-based one, use formula (20) and $tun_2$ to truncate, and reconstruct the term as $\textbf{R}_k^{(t)}$.

2: Preferences among pairs of samples for each $\textbf{R}_k^{(t)}$ are calculated through one mode of probability assignment P(.) using formulas (21) and (22).

3: A transition matrix is established and adapted using formula (23) and the tuning parameter $tun_1$.

4: The obtained transition matrix must be transposed and then used in a regular power iteration algorithm to calculate its left eigenvector (the stationary distribution).

5: The value of each element in a stationary distribution can be regarded as a Markov chain score of its corresponding samples. The $\textbf{R}_{agg}^{(t)}$ is the rank list of the Markov chain score from large to small, and the top $N$ samples with high Markov chain scores are collected as $\textbf{Q}^{(t)}$ values to query for labels.

**Output:** $\textbf{Q}^{(t)}$, which denotes the most $N$ valuable samples, is selected from $\textbf{U}^{(t)}$ in the $t$-th iteration.

Above all, the improved process of our RMQCAL can be defined as follows:

**Algorithm 5: Our RMQCAL process**

**Input:** The $L$ SQC, namely, $F_k^{(t)}$ ($k$=1: $L$), the unlabeled dataset $\textbf{U}^{(0)}$, and the number of samples selected in each iteration $N$.

**Repeat**

    **If** the number of iterations $t = 0$

        Step 0: randomly use one kind of experimental design-based SQC to select the first batch of unlabeled samples for labeling as $\textbf{Q}^{(0)}$, $\textbf{A}^{(0)} = \textbf{Q}^{(0)}$ and $\textbf{U}^{(1)} = \textbf{U}^{(0)} \setminus \textbf{Q}^{(0)}$

    **Else**

        Step 1: Obtain each pair of $\textbf{R}_k^{(t)}$ and $\textbf{S}_k^{(t)}$ using $F_k^{(t)}$ in $\textbf{U}^{(t)}$.

        Step 2: Using Algorithm 1 for $\textbf{S}_k^{(t)}$, obtain the weights of each $F_k^{(t)}$ in the $t$ th iteration $\textbf{S}_k^{(t)} \to \textbf{S}_k^{*(t)} \to \textbf{w}_k^{(t)}$

        Step 3: Choose one rank aggregation method (Algorithm 2, Algorithm 3, or Algorithm 4) to obtain the weighted aggregated rank list and select the sample with the top $N$ value as $\textbf{Q}^{(t)}$. Then, $N, \textbf{R}_k^{(t)}, \textbf{w}_k^{(t)} \to \textbf{Q}^{(t)}$

        Step 4: Request a label of $\textbf{Q}^{(t)}$ from the Oracle. Then, $\textbf{A}^{(t)} = \textbf{A}^{(t-1)} \cup \textbf{Q}^{(t)}$ and $\textbf{U}^{(t+1)} = \textbf{U}^{(t)} \setminus \textbf{Q}^{(t)}$.

    **End**

**Until:** a stopping criterion is applied or $|\textbf{U}^{(t)}| = 0$.

*2.4.4. Comparison of methods*

Note that, regarding levels of computational complexity, in the above weighted rank aggregation methods, the algorithm for solving the top-k problem in an unsorted array with *n* elements is unified, and the computational complexity of these methods is equal to $O(n)$. Then, on average, the computational complexity of Borda's and Bucklin voting are all $O(|U^{(t)}| \times L)$. Because we employ 'sample truncation', the computational complexity of the Markov Chain can be diminished to $O(N^{*3})$ from $O(|U^{(t)}|^3)$. Because the value of $N^*$ is far lower than that of $|U^{(t)}|$, the computational complexity of our improved Markov Chain method for MQCAL will be acceptable.

Borda's method is inferior to the others because in certain special cases, particularly when *L* is small or when one rank list $R_k^{(t)}$ is committee-based, using Borda's method can cause most unlabeled samples to have the same overall Borda score. In turn, the most valuable samples with the lowest overall Borda scores often cannot be selected. This situation does not occur when the Markov chain and Bucklin voting methods are applied. Considering the corresponding performance, relevant documents indicate that the Markov chain works better than the Borda and Bucklin voting methods when applied to traditional rank aggregation problems. For the RMQCAL problem, the rationality and validity of the above ranking aggregation method still need to be confirmed by multiple experiments, as described in Section 4.

## 3. Experiments

*3.1. Dataset description*

**Table 1**

The experimental datasets used

| | Positive &Negative | Feature Number | Sample Size | | Positive &Negative | Feature Number | Sample Size |
|---|---|---|---|---|---|---|---|
| Dataset for searching best combination of rank aggregation methods and SQC | | | | *Titato\** | 626:332 | 9 | 958 |
| *Wdbc\** | 212:357 | 30 | 569 | *Austra\** | 307:383 | 14 | 690 |
| Datasets for comparing RMQCAL with conventional AL with single criterion | | | | *LetterEF\** | 768:775 | 16 | 1543 |
| *F-O* | 1089:2554 | 51 | 3643 | *LetterIJ\** | 755:747 | 16 | 1502 |
| *Forest* | 159:195 | 27 | 354 | *LetterMN\** | 792:783 | 16 | 1575 |
| *Gesture* | 694:656 | 32 | 1350 | *LetterDP\** | 805:803 | 16 | 1608 |
| *Parkinson* | 520:520 | 26 | 1040 | *LetterUV\** | 813:764 | 16 | 1577 |
| *Seed* | 70:70 | 7 | 140 | Datasets for comparing RMQCAL with other state of the art MQCAL | | | |
| *Firm* | 4305:4300 | 16 | 8605 | *Mushroom+* | 4208:3916 | 22 | 8124 |
| Datasets for comparing RMQCAL with other state of the art MQCAL | | | | *EEG+* | 6723:8257 | 14 | 14980 |
| *Vehicle\** | 218:217 | 18 | 438 | *Mocap+* | 16265:15733 | 15 | 31998 |
| *Isolet\** | 300:300 | 617 | 600 | *Epilepsy+* | 2300:2300 | 178 | 4600 |

To evaluate the effectiveness of our RMQCAL, comparative experiments are conducted on 20 different binary classification problems through the UCI Repository drawn from the public website http://archive.ics.uci.edu/ml/. Each problem corresponds to one dataset, as shown in Table 1. The datasets *F-O, Forest, Gesture, Parkinson, Seed,* and *Firm* are used to compare the performance of the proposed

RMQCAL and single query criterion-based AL methods (while also serving as the components of our method). The datasets *Vehicle\**, *Isolet\**, *Titato\**, *Austra\**, *LetterEF\**, *LetterIJ\**, *LetterMN\**, *LetterDP\**, *LetterUV\** and *Wdbc\** are used for comparison between RMQCAL and the state-of-the-art AL and MQCAL methods, which are consistent with the datasets provided in [8]. Moreover, the remaining four datasets, *Mushroom[+]*, *EEG[+]*, *Mocap[+]*, and *Epilepsy[+]*, which contain more than 4000 samples, are used to validate the effectiveness of the proposed method on a large-scale dataset. It is worth mentioning that the dataset *Wdbc\** is also employed to search for the best combination of rank aggregation methods and SQC in RMQCAL.

Before an experiment is conducted, each dataset is normalized and randomly divided into two parts of equal size. One part is used as a test set, and the other is used as the unlabeled sample for AL methods. To ensure the reliability of the experimental results, most of the experiments listed below are run 10 times, and the average for each period is shown as the final performance result.

*3.2. Experimental setting*

All operations are executed using MATLAB R2014a software (Mathworks, Inc., Natick, MA, USA) installed on a PC with an Intel Core i3-2100 CPU (3.10 GHz) and 3 GB memory. Because the main purpose of AL is to effectively and efficiently establish a good learning model regardless of whether it improves its performance, this paper only applies to an SVM classifier with an RBF kernel—the same as was used in [8]—as the baseline against which comparisons to all approaches can be drawn. The SVM classifier is supported by LibSVM in http://www.csie.ntu.edu.tw/~cjlin/libsvm/. The source code of our RMQCAL and the control method also have been uploaded to GitHub. Any reader interested in this can download the code from https://github.com/wangtaoz/RMQCAL.git.

*3.3. Performance metrics*

For **Experiments B, C,** and **D,** two metrics (namely, accuracy and F1-measure) are used to evaluate the performance of approaches relative to those described in [8]. The F1-measure is a common metric described in formula (24):

$$F1 = \frac{2 \times Precision \times Recall}{Precision + Recall}. \tag{24}$$

For **Experiment E** and **Experiment F**, the area under the ROC curve (AUC) is added as an additional evaluation metric. In addition, paired t-tests conducted at the 95 percent significance level are introduced to reflect the difference between the two methods. For **Experiment A**, Kendal's tau distance [39] and Spearman's footrule distance [40] are introduced to evaluate rank aggregation effects as formula (25):

$$\begin{cases} dist_{Spearman} = \sum_{k=1}^{L} Spear(R_{agg}^{(t)}, R_k^{(t)}) = \sum_{k=1}^{L} \sum_{i=1}^{|U^{(t)}|} |R_{agg}^{(t)}(i) - R_k^{(t)}(i)| \\ dist_{Kendal} = \sum_{k=1}^{L} Kendal(R_{agg}^{(t)}, R_k^{(t)}) = \sum_{k=1}^{L} \sum_{i,j=1}^{|U^{(t)}|} I((R_{agg}^{(t)}(i) - R_{agg}^{(t)}(j))(R_k^{(t)}(i) - R_k^{(t)}(j)) < 0) \end{cases} \tag{25}$$

where $i \neq j \in [1, 2, \cdots, |U^{(t)}|]$, and $R_{agg}^{(t)}$ is the rank list after aggregation.

*3.4. Experimental goals*

The goal of **Experiment A** is to narrow the selection of rank aggregation methods using the contrastive experiment in a toy example. the selected methods will be used in **Experiment B,** which is designed to choose the best method from Borda's method, Bucklin voting and Markov Chain, and the best method will be used in the following experiments.

The goal of **Experiment C** is to find possible RMQCAL rules, including the results within various combinations of involved SQCs and the results for various numbers of involved SQCs. The optimal combination will be adopted in the rest of the experiments.

The goal of **Experiment D** is to confirm that the proposed RMQCAL can adequately combine multiple AL methods and achieve a higher performance than its components, i.e., individual AL methods that use a single-query criterion.

Different from **Experiment D**, **Experiment E** attempts to compare the proposed RMQCAL to the existing state-of-the-art MQCAL and other traditional AL methods.

**Experiment F** is used to validate the effectiveness of the proposed method on a large-scale dataset, when **Experiment G** is used to analyze and evaluate the algorithm efficiency of the proposed RMQCAL.

3.5. Experimental process

3.5.1. The selection of the most appropriate rank aggregation methods for MQCAL

To preliminarily narrow the selection of rank aggregation methods, in **Experiment A**, a faux example is presented as an input list with seven ranking lists to illustrate the performance differences of the candidate rank aggregation methods; which are confirmed as available in the above article including the Borda methods within different Borda score expressions (i.e., median, p-norm, minimum and geometric mean, as described in formula (19)), Bucklin voting, and Markov chain methods within several weighted transition probability expressions (i.e., MC1, MC2, MC3, as described in formula (21)). Moreover, in this and the following experiments, the tuning parameter of the Markov chain method and the Borda parameter are set as 0.05 and 1, respectively.

Table 2

Comparison of ranking aggregation methods used in a toy example

| Sample | Input Lists | | | | | | | Borda Method | | | | | Markov Chain | | |
|---|---|---|---|---|---|---|---|---|---|---|---|---|---|---|---|
| | L1 | L2 | L3 | L4 | L5 | L6 | L7 | Min | Med | P-norm | Geo | Buc | MC1 | MC2 | MC3 |
| Sample1 | 8 | 6 | 3 | 3 | 7 | 4 | 1 | 1 | 2 | 3 | 3 | 4 | 8 | 3 | 3 |
| Sample2 | 10 | 8 | 2 | 2 | 9 | 5 | 1 | 6 | 4 | 5 | 4 | 7 | 7 | 4 | 7 |
| Sample3 | 2 | 2 | 1 | 1 | 5 | 3 | 1 | 2 | 1 | 1 | 1 | 1 | 1 | 1 | 1 |
| Sample4 | 9 | 1 | 4 | 4 | 3 | 2 | 1 | 3 | 3 | 2 | 2 | 3 | 3 | 2 | 2 |
| Sample5 | 5 | 7 | 8 | 10 | 10 | 7 | 5 | 9 | 7 | 9 | 9 | 9 | 9 | 8 | 10 |
| Sample6 | 7 | 10 | 8 | 8 | 1 | 1 | 5 | 4 | 8 | 7 | 5 | 5 | 5 | 9 | 5 |
| Sample7 | 6 | 5 | 8 | 9 | 8 | 10 | 5 | 10 | 10 | 10 | 10 | 10 | 10 | 10 | 9 |
| Sample8 | 1 | 9 | 6 | 7 | 2 | 8 | 5 | 5 | 9 | 6 | 6 | 4 | 4 | 6 | 5 |
| Sample9 | 4 | 4 | 5 | 6 | 6 | 9 | 5 | 8 | 6 | 8 | 8 | 6 | 6 | 7 | 8 |
| Sample10 | 3 | 3 | 7 | 5 | 4 | 6 | 5 | 7 | 5 | 4 | 7 | 2 | 2 | 5 | 4 |
| Kendall's tau distance | | | | | | | | 93 | 83 | **81** | 84 | **85** | 99 | **79** | 84 |
| Spearman's footrule distance | | | | | | | | 178 | 156 | **156** | 166 | **152** | 172 | **154** | 158 |

From the results in Table 2, it can be observed that the MC2 method, the Bucklin method and the Borda method based on p-norm perform better with Kendall's tau distance and Spearman's distance values of (79,154), (85,152) and (81,156), respectively. Therefore, for **Experiment B**, we only consider these three rank aggregation methods as MQCAL integration criteria strategies and disregard the others.

For **Experiment B**, a complete RMQCAL process is implemented for dataset *Wdbc\**. In this phase, the controlled experiment is designed to evaluate the effects of various rank aggregation methods on our MQCAL process. The SQC are fixed. The Borda (with p-norm and *p* equal to 1), Bucklin, and improved Markov chain (MC2) methods are used to construct a ranking aggregation process, and their accuracy levels and F1-measures (X-axis) with different labeling costs (Y-axis), which are indicated as the percentage of the sample designated for label selection, are compared in Fig. 6.

The results of **Experiment B** are shown in Fig. 6. Beyond the three rank aggregation methods used for integration, we find that the enhanced effects of our RMQCAL method based on the Markov chain approach with an MC2 weighted transition probability expression are the most easily detectable; the Borda approach ranks second and the Bucklin voting method is not satisfactory. This final result is likely attributable to the fact that the Bucklin voting method often positions samples with the highest median ratings at the top of a rank list after aggregation, which is not suitable for AL problems. Thus, subsequent experiments use the Markov chain approach with an MC2 weighted transition probability expression.

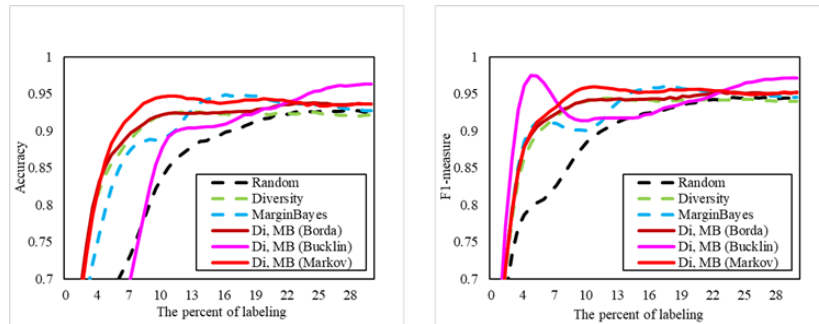

**Fig. 6.** Performance of RMQCAL realized using the Borda, Bucklin voting and Markov chain methods

*3.5.2. RMQCAL experiments with various SQCs combinations*

` After determining the best ranking aggregation method for the proposed RMQCAL method in **Experiments A and B**, **Experiment C** involves a series of comparative experiments that are used to reflect the performance of RMQCAL in dataset *Wdbc\** when applying various combinations of SQCs. The SQCs used include the following: Diversity (DI), Margin in RBF-SVM (MR), Margin in Bayes (MB), QBC and TED. The performance curves of each case, including their accuracies and F1-measures (X-axis) with different labeling costs (Y-axis), are presented in Fig. 7 to Fig. 15.

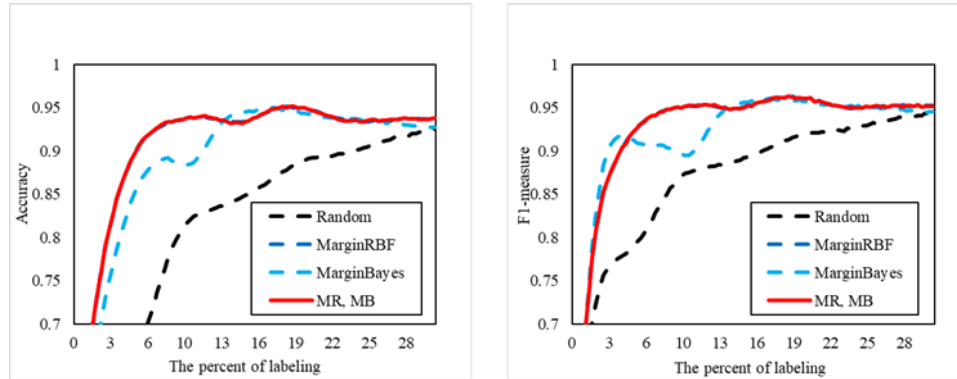

**Fig. 7.** Performance of RMQCAL with two certainty-based SQCs

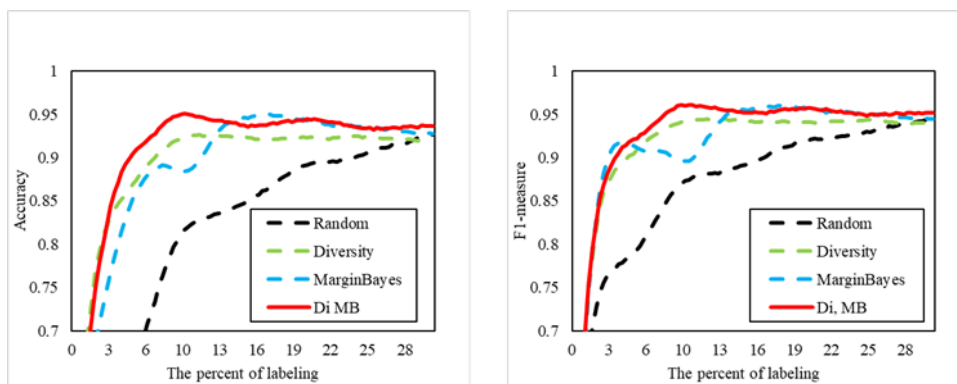

**Fig. 8.** Performance of RMQCAL with certainty-based and connection-based SQCs

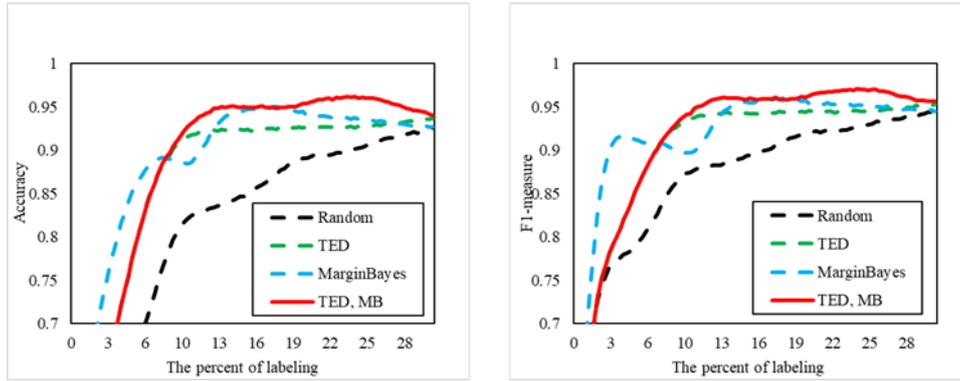

**Fig. 9.** Performance of RMQCAL with SQCs based on certainty and experimental design

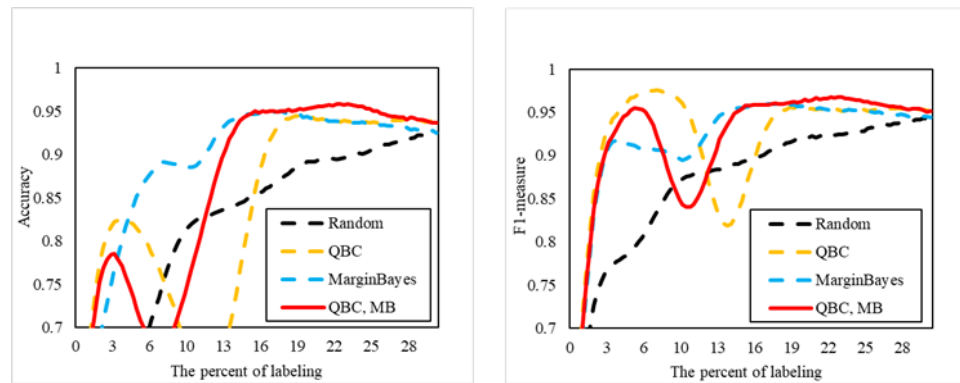

**Fig. 10.** Performance of RMQCAL with certainty-based and committee-based SQCs

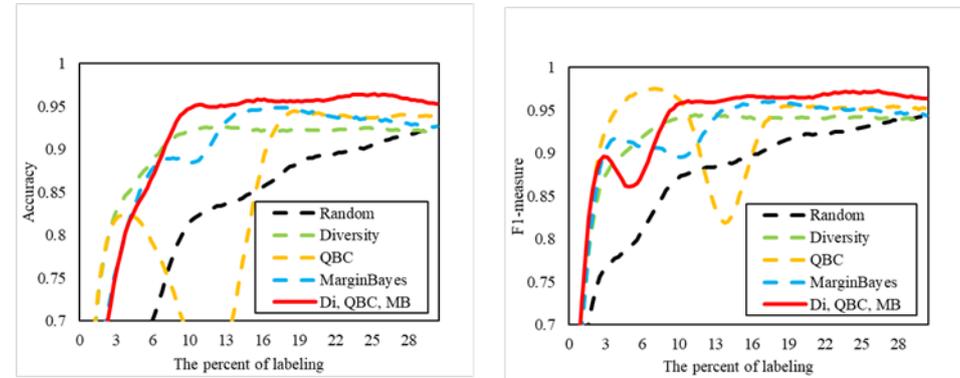

**Fig. 11.** Performance of RMQCAL with SQCs based on certainty, committee, connection

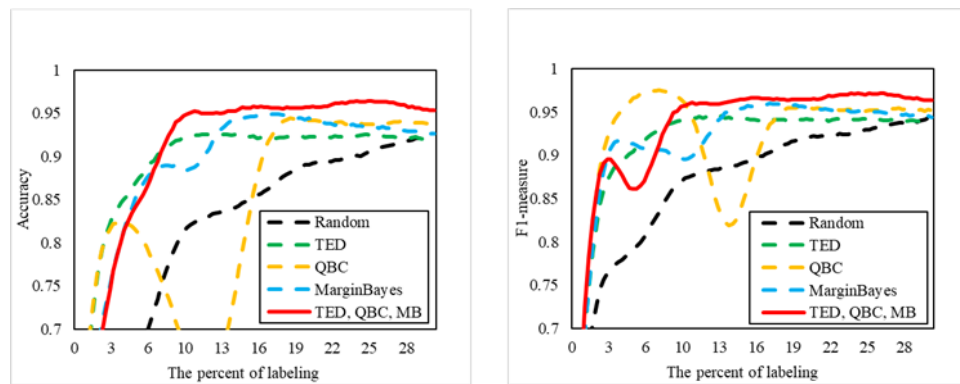

**Fig. 12.** Performance of RMQCAL with SQCs based on certainty, committee, experimental design

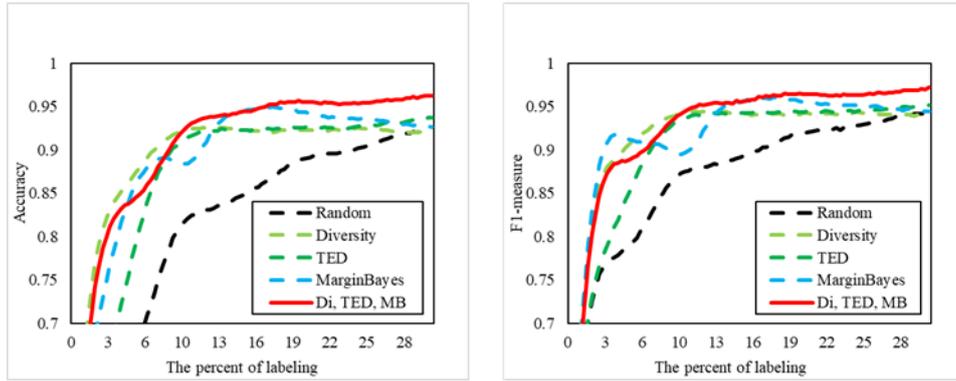

**Fig. 13.** Performance of RMQCAL with SQCs based on certainty, connection and experimental design

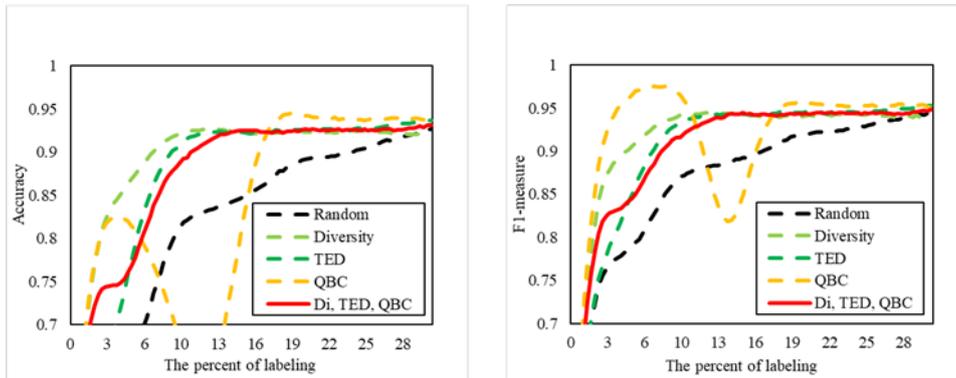

**Fig. 14.** Performance of RMQCAL with SQCs based on committee, connection and experimental design

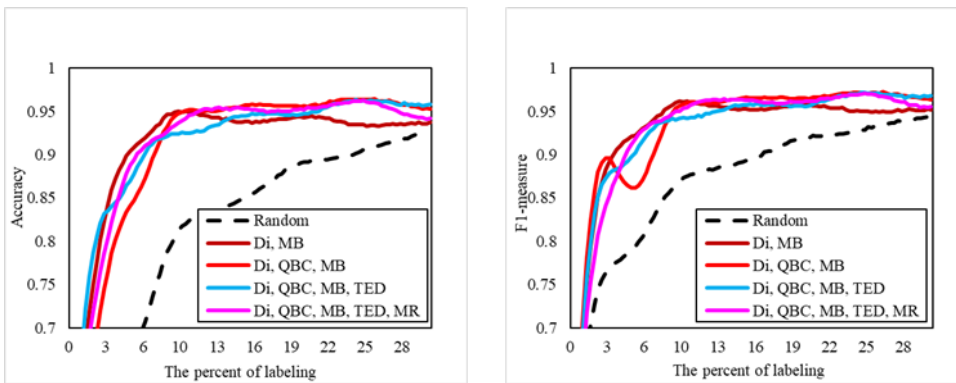

**Fig. 15.** Performance of RMQCAL with several SQCs

**Experiment C** helps explain several of the problems. First, the proposed MQCAL does have high scalability, which would make it able to offer a variety of criteria combinations with less algorithm modification. Second, the RMQCAL with the combination of multiple SQC usually (not always) performs better than its components: i.e., an AL with a single criterion. However, an inappropriate combination may lead to no definable benefit, as shown, for example, in Fig. 7, Fig. 10 and Fig. 14. This article considers that the failure of the dynamic weighting process is the primary explanation for these exceptions. To prove this, we specifically record weight changes in each involved single criterion under the experimental conditions used for Fig. 8, Fig. 9, Fig. 7 and Fig. 10 as well as Fig. 16, Fig. 17, Fig. 18 and Fig. 19, where coordinate-axis X indicates the labeling cost and coordinate-axis Y indicates the value of self-adaptive weights. Under ideal conditions, the weight changes of each single criterion involved in the AL process should be dynamic and should satisfactorily reflect the contribution of each criterion. In the successful cases shown in Fig. 8 and Fig. 9, both of their weight changes in Fig. 16 and Fig. 17 show a gradual decline in weight for the representativeness measure-based SQC, whereas the weight of the informativeness measure-based SQC rises continuously. Combined with 'the timeliness of AL' described above, such weight changes are just what we need. Conversely, as to the ineffective cases, the involved two SQC in Fig. 7 belong to the same category (Certainty), and one (MR) is always better than the others (MB), causing our RMQCAL to assign a higher weight to the MR from beginning to end (as shown in Fig. 18). The combination that involved committee-based SQC in Fig. 10 also seems not to perform well, and we believe that this is because the involved SQC will be equally weighted in our designed weighting calculation step only if there are only two SQC and one of them is QBC (as shown in Fig. 19). Similarly, it is not surprising that the case shown in Fig. 14 does not perform well not only because it involves committee-based SQC but also because another two involved SQC also belong to same type of AL (i.e., representativeness measure-based).

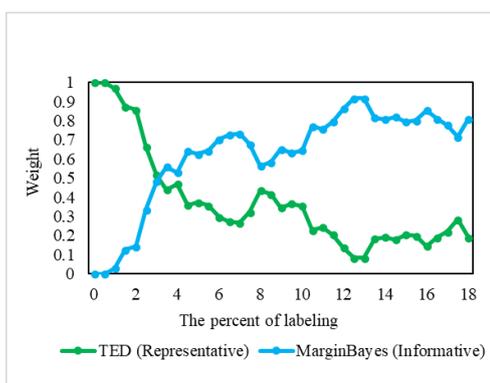
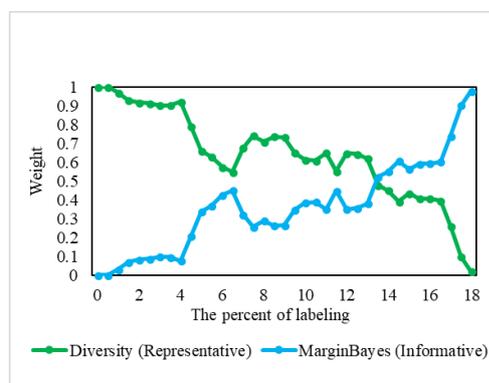

**Fig. 16.** Weight variations of TED and MB    **Fig. 17.** Weight variations of DI and MB

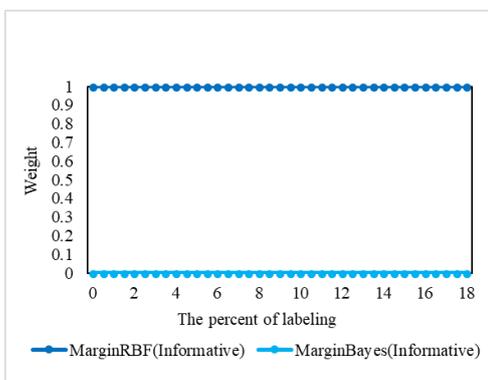
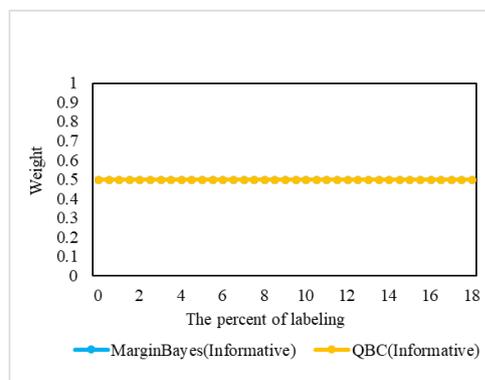

**Fig. 18.** Weight variations of MR and MB    **Fig. 19.** Weight variations of QBC and MB

Moreover, Fig. 15 indicates that with the increasing number of SQC from more AL methods, the performance of RMQCAL improves and becomes more stable, although the room for continued improvement gets smaller. On the other hand, as shown in Fig. 15, the potential predominance of RMQCAL in practical applications can also be partially supported if we regard these involved AL methods as the candidate methods.

In sum, several rules exist by which SQC are selected for combination. (1) The SQC do not have to be numerous. In general, three SQC are sufficient. (2) The involved SQC preferably belong to different types of AL. At least one of them is certainty-based, and another one is representativeness measure-based. Committee-based SQC should be used in conjunction with two or more different types of SQC. (3)

Certainty-based SQC can promote the performance of our RMQCAL in the middle stages of the process. Representativeness measure-based SQC can enhance AL performance in the early stages. QBC can smooth the performance curve. According to the above rules and results shown in Fig. 15, we recommend employing Diversity, Margin and QBC as three involved SQC in the proposed RMQCAL. These will also be used in all tests shown below. In our view, this combination is both typical and adequate.

*3.5.3. Comparisons between RMQCAL and single-query criterion-based AL methods*

For **Experiment D**, additional datasets (i.e., *F-O*, *Forest*, *Gesture*, *Parkinson*, *Seed*, and *Firm*) are introduced to demonstrate that the proposed RMQCAL can adequately combine multiple AL methods and achieve a higher performance than its individual components (i.e., the AL methods with a single-query criterion). We specifically use *Diversity* and *Margin* in RBF-SVM and *QBC* in our RMQCAL. Although this combination may not be the best, it is the most representative because it includes three typical types of SQCs. The experiment for each dataset must be repeated 10 times; Table 3 shows the average accuracy of each method for 5, 10, 15, 20, 25, 30 percent of the unlabeled data. For each case, the best results are highlighted in boldface. The win/tie/loss counts of RMQCAL versus its components (i.e., the conventional single-query criterion-based AL methods) are also presented in Table 4.

Table 3

Accuracy of our RMQCAL method when applied to 6 UCI datasets

| % | Algorithms | Firm | F-O | Fore | Gest | Park | Seed | % | Algorithms | Firm | F-O | Fore | Gest | Park | Seed |
|---|---|---|---|---|---|---|---|---|---|---|---|---|---|---|---|
|  | Random | 0.522 | **0.701** | 0.549 | 0.486 | **0.500** | 0.750 |  | Random | 0.952 | 0.701 | 0.715 | 0.582 | 0.502 | 0.921 |
|  | Di | 0.573 | **0.701** | **0.705** | 0.486 | **0.500** | 0.908 |  | Di | 0.955 | **0.703** | 0.808 | 0.486 | 0.500 | 0.934 |
| 5 | QBC | 0.500 | 0.299 | 0.451 | 0.486 | **0.500** | 0.921 | 20 | QBC | 0.956 | 0.299 | 0.451 | 0.514 | 0.500 | 0.934 |
|  | MR | 0.573 | **0.701** | 0.451 | **0.731** | **0.500** | 0.921 |  | MR | **0.971** | **0.703** | 0.891 | 0.799 | 0.526 | 0.934 |
|  | **RMQCAL** | **0.745** | **0.701** | 0.451 | 0.514 | **0.500** | **0.934** |  | **RMQCAL** | 0.967 | 0.699 | **0.896** | **0.849** | **0.597** | **0.935** |
|  | Random | 0.938 | 0.701 | 0.549 | 0.517 | 0.500 | 0.921 |  | Random | 0.958 | 0.701 | 0.767 | 0.634 | 0.500 | **0.934** |
|  | Di | 0.903 | 0.435 | 0.803 | 0.486 | 0.500 | **0.934** |  | Di | 0.960 | 0.702 | 0.865 | 0.486 | 0.500 | **0.934** |
| 10 | QBC | 0.835 | 0.299 | 0.451 | 0.514 | 0.500 | 0.868 | 25 | QBC | 0.965 | 0.299 | 0.451 | 0.514 | 0.500 | **0.934** |
|  | MR | 0.954 | **0.710** | **0.865** | 0.712 | 0.500 | 0.921 |  | MR | 0.970 | 0.702 | **0.907** | 0.830 | 0.574 | **0.934** |
|  | **RMQCAL** | **0.955** | 0.685 | 0.855 | **0.837** | **0.504** | **0.934** |  | **RMQCAL** | **0.970** | **0.704** | 0.896 | **0.858** | **0.600** | **0.934** |
|  | Random | 0.951 | 0.701 | 0.777 | 0.529 | 0.500 | 0.908 |  | **Random** | 0.959 | 0.701 | 0.829 | 0.657 | 0.509 | **0.934** |
|  | Di | 0.941 | **0.714** | 0.813 | 0.486 | 0.500 | **0.934** |  | **Di** | 0.961 | 0.701 | 0.896 | 0.486 | 0.500 | **0.934** |
| 15 | QBC | 0.924 | 0.299 | 0.451 | 0.514 | 0.500 | **0.934** | 30 | **QBC** | 0.972 | 0.299 | 0.451 | 0.514 | 0.500 | **0.934** |
|  | MR | 0.957 | 0.704 | 0.876 | 0.813 | 0.502 | 0.908 |  | **MR** | 0.971 | 0.701 | 0.782 | 0.722 | 0.597 | 0.921 |
|  | **RMQCAL** | **0.965** | 0.674 | **0.881** | **0.845** | **0.516** | 0.921 |  | **RMQCAL** | **0.974** | **0.711** | **0.912** | **0.853** | **0.618** | **0.934** |

Table 4

Win/Tie/Loss Counts of RMQCAL versus its components

| | Percentage of the unlabeled samples | | | | | | |
|---|---|---|---|---|---|---|---|
| | 5% | 10% | 15% | 20% | 25% | 30% | IN ALL |
| Algorithms | WIN/TIE/LOSS | WIN/TIE/LOSS | WIN/TIE/LOSS | WIN/TIE/LOSS | WIN/TIE/LOSS | WIN/TIE/LOSS | WIN/TIE/LOSS |
| Random | 4/2/0 | 5/0/1 | 5/0/1 | 5/0/1 | 5/1/0 | 5/1/0 | 29/4/3 |
| Di | 3/2/1 | 5/1/0 | 4/0/2 | 5/0/1 | 5/1/0 | 5/1/0 | 27/5/4 |
| QBC | 5/1/0 | 6/0/0 | 5/0/1 | 6/0/0 | 5/1/0 | 5/1/0 | 32/3/1 |
| MR | 2/2/2 | 4/0/2 | 4/0/2 | 4/0/2 | 3/2/1 | 6/0/0 | 23/4/9 |
| IN ALL | 14/7/3 | 20/1/3 | 18/0/6 | 20/0/4 | 18/5/1 | 21/3/0 | 111/16/17 |

The results shown in Table 3 and Table 4 illustrate that RMQCAL wins or ties accounts for nearly 90 percent of the total, further confirming that through RMQCAL, the combination of multiple appropriate AL methods can help the user obtain a better classification model with less labeling cost than can be achieved from the use of any individual AL method.

*3.5.4. The comparisons between RMQCAL and state of the art AL methods in medium scale data sets*

**Experiment E** is the focus of our experiments, which compare our research results with state-of-the-art AL methods and MQCAL methods of various SQC and integration criteria strategies. The controlled methods include MARGIN [41], CLUSTER [10], IDE [42], DUAL [7], QUIRE [8], SMQCAL [23] (serial-form), PMQCAL [23] (parallel-form), CSAL [21] (based on criteria selection) and MCDMAL [9] (based on multicriteria decision-making). The first two methods are classic AL methods based on an SQC that uses informativeness and representativeness measures. The other seven methods are well-developed approaches of other forms of MQCAL. To ensure the validity of the comparative experiments and avoid the effects of other factors, we reproduce an experimental environment that is exactly the same as that described in another paper [8]. The same 10 datasets (i.e., *Wdbc\*, Vehicle\*, Isolet\*, Titato\*, Austra\*, LetterDP\*, LetterEF\*, LetterIJ\*, LetterMN\** and *LetterUV\**) are used with the same experimental parameters provided in the corresponding literature to determine whether the proposed RMQCAL is competitive with existing MQCAL methods.

The SQCs involved in our RMQCAL method include DIVERSITY and MARGIN in RBF-SVM and QBC, which may be not optimal but nevertheless represents the most typical combination involving three kinds of single-criterion AL methods. With respect to the integration strategy, SQC employs an improved rank aggregation method based on Markov chain.

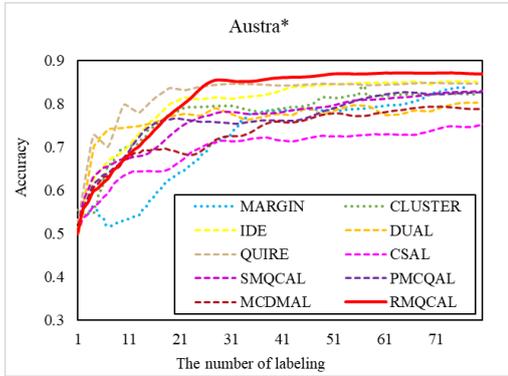

a: RMQCAL in *Austra\**

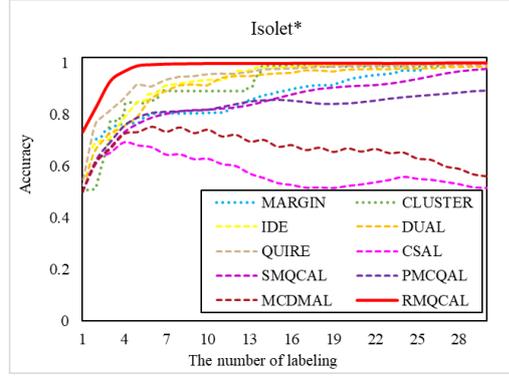

b: RMQCAL in *Isolet\**

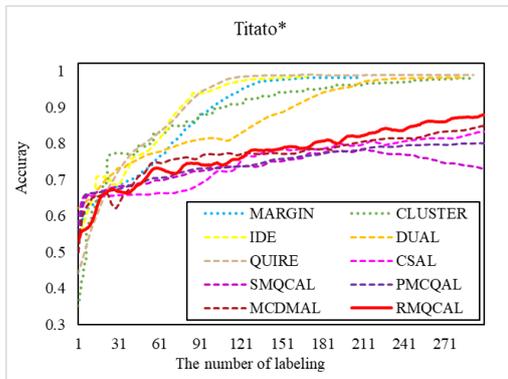

c: RMQCAL in *Titato\**

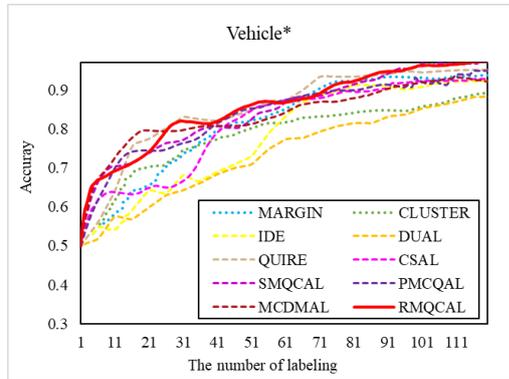

d: RMQCAL in *Vehicle\**

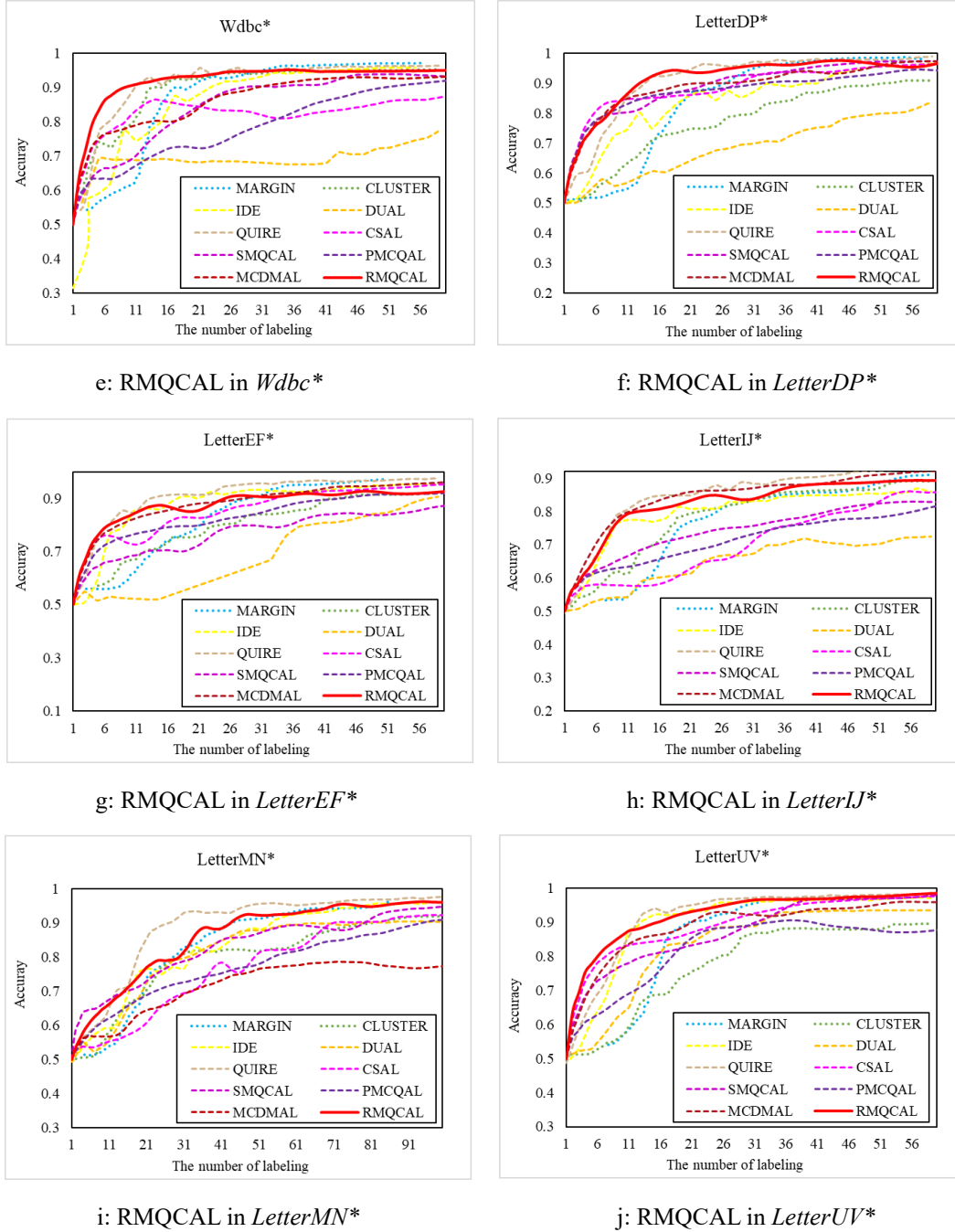

**Fig. 20.** Comparisons of the accuracy of the 10 datasets

For each dataset, a corresponding experiment is repeated 10 times, and performance curves measuring accuracy (X-axis) against labeling costs (Y-axis) for the methods are shown in Fig. 20. In addition to the average AUC (MEAN) of each method, as calculated using 5, 10, 20, 30, 40 percent of the data unlabeled, we also record their standard deviations (SDs) in Table 5, Table 6 and Table 7. The best result and its performance are recorded in bold based on paired t-tests conducted at the 95-percent significance level. A more detailed comparison between RMQCAL and another MQCAL method is shown in Table 8.

From the comparative results shown in Table 8 and the 10 graphs shown in Fig. 20, the results (i.e., RANDOM, CLUSTER, and MARGIN) of the proposed RMQCAL method were significantly better than those of the AL methods based on a single criterion. Regarding MQCAL with various integration criteria strategies (i.e., IDE, DUAL, SMQCAL, PMQCAL, CSAL and MCDMAL), although we made our best effort to tune their related weight parameters or directly use the recommended values provided in corresponding studies, these methods are still sometimes inferior to conventional AL methods based on a single criterion. We suggest that the relatively low uniformity and generality of these methods, caused by excessive dependence on empirical parameters and the tuning process, are the reasons for their suboptimal performance. From Table 8, the proposed method does not seem to outperform QUIRE, but we believe that the second-place performance of the RMQCAL method relative to the other methods is acceptable for the following reasons. First, QUIRE is directly derivable from the SVM framework, and its representativeness and informativeness calculations are highly consistent. The main research content in QUIRE is the design of SQC for both representativeness and informativeness measures rather than how to combine them together. Conversely, similar to other MQCAL methods, RMQCAL focuses on the design of the integration criteria strategy, and the involved SQCs could emerge from existing ordinary methods, which could include the QUIRE method, as the tests in the appendix prove that the combination of QUIRE and other SQCs via RMQCAL could be better than using QUIRE alone. Second, the RMQCAL method can produce an effect similar to a well-designed AL method by combining several existing ordinary SQCs. Third, compared with QUIRE, RMQCAL also has higher efficiency, which has been discussed in **Experiment G**. In addition, as shown in the results in the appendix, the performance of QUIRE is very sensitive to empirical parameters; if a parameter is inappropriate, the performance of the method could be worse than that of RMQCAL. Meanwhile, setting optimal parameters requires extra labeling to build the validation set.

Table 5

Comparison of the AUC values of the 10 datasets (1)

| Database | Algorithms | Percentage of the unlabeled samples | | | | |
| --- | --- | --- | --- | --- | --- | --- |
| | | 5% Mean±SD | 10% Mean±SD | 20% Mean±SD | 30% Mean±SD | 40% Mean±SD |
| Austra | RANDOM | 0.868±0.027 | 0.894±0.022 | 0.897±0.023 | 0.901±0.022 | 0.909±0.015 |
| | MARGIN | 0.751±0.137 | 0.838±0.119 | 0.885±0.043 | 0.909±0.010 | 0.911±0.012 |
| | CLUSTER | 0.877±0.045 | 0.888±0.029 | 0.894±0.015 | 0.896±0.015 | 0.903±0.014 |
| | IDE | **0.858±0.101** | **0.885±0.058** | 0.902±0.012 | 0.912±0.008 | 0.913±0.009 |
| | DUAL | 0.866±0.037 | 0.878±0.036 | 0.875±0.018 | 0.876±0.016 | 0.879±0.013 |
| | QUIRE | 0.887±0.014 | 0.901±0.010 | 0.906±0.016 | 0.912±0.009 | 0.914±0.009 |
| | CSAL | 0.820±0.055 | 0.840±0.054 | 0.856±0.039 | 0.866±0.023 | 0.860±0.024 |
| | SMCDMAL | 0.850±0.019 | 0.851±0.023 | 0.879±0.016 | 0.886±0.011 | 0.896±0.012 |
| | PMCQAL | 0.835±0.058 | 0.849±0.047 | 0.879±0.022 | 0.894±0.020 | 0.907±0.016 |
| | MCDMAL | 0.843±0.045 | 0.839±0.037 | 0.836±0.030 | 0.855±0.023 | 0.863±0.019 |
| | **RMQCAL** | **0.916±0.009** | **0.921±0.008** | **0.926±0.007** | **0.927±0.007** | **0.929±0.007** |
| Isolet | RANDOM | 0.995±0.006 | 0.998±0.002 | 0.999±0.001 | 1.000±0.000 | 1.000±0.000 |
| | MARGIN | 0.965±0.052 | 0.999±0.001 | **1.000±0.000** | 1.000±0.000 | 1.000±0.000 |
| | CLUSTER | 0.998±0.002 | **0.999±0.002** | 1.000±0.000 | 1.000±0.000 | 1.000±0.000 |
| | IDE | **0.998±0.003** | **0.999±0.002** | 0.999±0.001 | 1.000±0.001 | 1.000±0.000 |
| | DUAL | 0.993±0.008 | 0.999±0.001 | 0.999±0.001 | 1.000±0.000 | 1.000±0.001 |
| | QUIRE | 0.997±0.002 | 0.999±0.001 | 0.999±0.001 | 1.000±0.000 | 1.000±0.001 |
| | CSAL | 0.948±0.047 | 0.934±0.039 | 0.895±0.059 | 0.935±0.059 | 0.961±0.056 |
| | SMCDMAL | **1.000±0.001** | **1.000±0.001** | **1.000±0.000** | **1.000±0.000** | **1.000±0.000** |
| | PMCQAL | **1.000±0.001** | **1.000±0.000** | **1.000±0.000** | **1.000±0.000** | **1.000±0.000** |
| | MCDMAL | 0.998±0.003 | **0.991±0.018** | 0.965±0.047 | 0.928±0.097 | 0.935±0.085 |
| | **RMQCAL** | **1.000±0.001** | **1.000±0.001** | **1.000±0.001** | **1.000±0.000** | **1.000±0.001** |

**Table 6**

Comparison of the AUC values of the 10 datasets (2)

| Database | Algorithms | Percentage of the unlabeled samples | | | | |
|---|---|---|---|---|---|---|
| | | 5%<br>Mean±SD | 10%<br>Mean±SD | 20%<br>Mean±SD | 30%<br>Mean±SD | 40%<br>Mean±SD |
| **Titato** | RANDOM | **0.762±0.033** | 0.861±0.031 | 0.954±0.023 | 0.979±0.011 | 0.991±0.007 |
| | MARGIN | 0.645±0.096 | 0.753±0.078 | 0.946±0.043 | 0.998±0.001 | **1.000±0.000** |
| | CLUSTER | **0.717±0.087** | 0.806±0.054 | 0.908±0.031 | 0.971±0.021 | 0.989±0.010 |
| | IDE | **0.735±0.040** | **0.906±0.029** | **0.996±0.003** | **0.999±0.001** | **1.000±0.001** |
| | DUAL | **0.708±0.069** | 0.782±0.064 | 0.900±0.027 | 0.981±0.012 | 0.995±0.006 |
| | QUIRE | **0.736±0.037** | 0.861±0.025 | 0.991±0.004 | **0.999±0.001** | **1.000±0.000** |
| | CSAL | 0.642±0.032 | 0.673±0.047 | 0.745±0.032 | 0.824±0.030 | 0.852±0.030 |
| | SMCDMAL | **0.690±0.046** | 0.740±0.033 | 0.824±0.018 | 0.862±0.010 | 0.869±0.021 |
| | PMCQAL | 0.671±0.038 | 0.745±0.044 | 0.805±0.030 | 0.861±0.029 | 0.887±0.030 |
| | MCDMAL | **0.719±0.031** | 0.792±0.031 | 0.823±0.012 | 0.845±0.016 | 0.863±0.023 |
| | **RMQCAL** | **0.732±0.074** | 0.813±0.050 | 0.860±0.059 | 0.933±0.024 | 0.964±0.019 |
| **Vehicle** | RANDOM | 0.818±0.064 | **0.864±0.039** | 0.925±0.032 | 0.949±0.026 | 0.968±0.016 |
| | MARGIN | 0.693±0.078 | 0.828±0.077 | 0.883±0.105 | **0.981±0.014** | **0.993±0.005** |
| | CLUSTER | 0.771±0.088 | 0.845±0.056 | 0.927±0.022 | 0.955±0.018 | 0.973±0.010 |
| | IDE | 0.731±0.141 | **0.849±0.106** | 0.878±0.093 | 0.957±0.037 | 0.977±0.010 |
| | DUAL | 0.680±0.074 | 0.706±0.114 | 0.817±0.061 | 0.875±0.035 | 0.908±0.035 |
| | QUIRE | 0.750±0.137 | **0.912±0.024** | **0.956±0.025** | **0.985±0.007** | 0.989±0.006 |
| | CSAL | 0.795±0.093 | 0.826±0.072 | 0.880±0.058 | 0.938±0.039 | 0.965±0.017 |
| | SMCDMAL | 0.825±0.044 | **0.888±0.032** | 0.931±0.020 | 0.954±0.019 | 0.972±0.019 |
| | PMCQAL | 0.872±0.036 | **0.895±0.038** | 0.934±0.029 | 0.967±0.016 | 0.983±0.011 |
| | MCDMAL | **0.884±0.037** | **0.903±0.031** | 0.927±0.028 | 0.950±0.021 | 0.967±0.014 |
| | **RMQCAL** | 0.806±0.077 | **0.905±0.053** | **0.958±0.016** | **0.986±0.014** | **0.996±0.006** |
| **Wdbc** | RANDOM | 0.984±0.006 | 0.986±0.005 | 0.990±0.004 | 0.991±0.004 | 0.991±0.004 |
| | MARGIN | **0.967±0.038** | 0.990±0.002 | **0.993±0.003** | 0.993±0.003 | 0.993±0.003 |
| | CLUSTER | 0.981±0.007 | 0.987±0.004 | 0.991±0.003 | 0.992±0.003 | 0.992±0.003 |
| | IDE | 0.983±0.006 | 0.984±0.008 | 0.990±0.004 | 0.992±0.003 | 0.993±0.003 |
| | DUAL | 0.955±0.025 | 0.964±0.016 | 0.972±0.015 | 0.988±0.009 | 0.992±0.003 |
| | QUIRE | 0.985±0.006 | **0.990±0.004** | **0.993±0.003** | 0.993±0.003 | 0.993±0.003 |
| | CSAL | 0.967±0.017 | 0.972±0.012 | 0.975±0.010 | 0.978±0.007 | 0.985±0.006 |
| | SMCDMAL | 0.954±0.030 | 0.968±0.015 | 0.982±0.005 | 0.984±0.007 | 0.985±0.004 |
| | PMCQAL | 0.923±0.063 | 0.954±0.027 | 0.968±0.012 | 0.984±0.007 | 0.985±0.005 |
| | MCDMAL | 0.934±0.065 | 0.959±0.033 | 0.981±0.006 | 0.982±0.006 | 0.983±0.006 |
| | **RMQCAL** | **0.993±0.005** | **0.993±0.003** | **0.995±0.003** | **0.996±0.002** | **0.997±0.003** |
| **LetterDP** | RANDOM | 0.990±0.004 | 0.995±0.002 | 0.997±0.002 | 0.998±0.001 | 0.998±0.001 |
| | MARGIN | 0.994±0.005 | **0.999±0.001** | 0.999±0.000 | 0.999±0.001 | 0.999±0.001 |
| | CLUSTER | 0.988±0.008 | 0.995±0.004 | 0.997±0.002 | 0.998±0.001 | 0.999±0.001 |
| | IDE | 0.992±0.006 | 0.997±0.002 | 0.998±0.001 | 0.999±0.001 | 0.999±0.001 |
| | DUAL | 0.978±0.005 | 0.986±0.001 | 0.988±0.004 | 0.990±0.004 | 0.996±0.001 |
| | QUIRE | **0.998±0.001** | **0.999±0.001** | 0.999±0.001 | 0.999±0.001 | 0.999±0.001 |
| | CSAL | 0.987±0.008 | 0.995±0.004 | **0.999±0.001** | 1.000±0.001 | 1.000±0.000 |
| | SMCDMAL | 0.992±0.005 | **0.998±0.003** | **1.000±0.000** | 1.000±0.000 | 1.000±0.000 |
| | PMCQAL | 0.988±0.007 | **0.988±0.019** | **1.000±0.001** | **1.000±0.000** | **1.000±0.000** |
| | MCDMAL | 0.993±0.003 | **0.997±0.003** | 0.999±0.001 | 1.000±0.000 | **1.000±0.000** |
| | **RMQCAL** | **0.997±0.001** | **0.999±0.001** | **0.999±0.001** | **1.000±0.001** | **1.000±0.001** |

Table 7

Comparison of the AUC values of the 10 datasets (3)

| Database | Algorithms | Percentage of the unlabeled samples | | | | |
|---|---|---|---|---|---|---|
| | | 5%<br>Mean±SD | 10%<br>Mean±SD | 20%<br>Mean±SD | 30%<br>Mean±SD | 40%<br>Mean±SD |
| **LetterEF** | RANDOM | **0.977±0.020** | 0.988±0.009 | 0.994±0.002 | 0.997±0.002 | 0.998±0.001 |
| | MARGIN | **0.987±0.008** | **0.999±0.001** | **1.000±0.000** | **1.000±0.000** | **1.000±0.000** |
| | CLUSTER | **0.975±0.016** | 0.991±0.003 | 0.997±0.004 | 0.999±0.001 | **1.000±0.000** |
| | IDE | **0.977±0.014** | 0.995±0.003 | 0.999±0.000 | 0.999±0.000 | 0.999±0.000 |
| | DUAL | **0.976±0.011** | 0.993±0.003 | 0.996±0.002 | 0.996±0.002 | 0.996±0.002 |
| | QUIRE | **0.988±0.009** | **0.999±0.000** | **1.000±0.000** | **1.000±0.000** | **1.000±0.000** |
| | CSAL | **0.981±0.008** | 0.991±0.005 | 0.999±0.001 | 0.999±0.000 | **1.000±0.000** |
| | SMCDMAL | 0.964±0.012 | 0.983±0.009 | 0.995±0.006 | 0.999±0.001 | 0.999±0.001 |
| | PMCQAL | **0.970±0.017** | 0.984±0.008 | 0.993±0.006 | 0.999±0.001 | 1.000±0.000 |
| | MCDMAL | **0.981±0.014** | 0.993±0.007 | 0.998±0.001 | 0.999±0.001 | **1.000±0.000** |
| | **RMQCAL** | **0.982±0.007** | 0.988±0.003 | 0.999±0.001 | 0.999±0.000 | 0.999±0.000 |
| **LetterIJ** | RANDOM | **0.943±0.025** | **0.966±0.017** | **0.980±0.004** | 0.983±0.005 | 0.985±0.005 |
| | MARGIN | 0.882±0.096 | **0.960±0.027** | **0.986±0.005** | **0.989±0.006** | **0.991±0.004** |
| | CLUSTER | **0.952±0.022** | **0.961±0.017** | **0.976±0.008** | 0.985±0.007 | **0.987±0.006** |
| | IDE | 0.934±0.030 | **0.969±0.011** | **0.979±0.006** | 0.980±0.006 | 0.982±0.008 |
| | DUAL | 0.819±0.120 | 0.897±0.058 | 0.934±0.030 | 0.954±0.017 | 0.959±0.014 |
| | QUIRE | **0.951±0.023** | **0.963±0.013** | **0.976±0.011** | **0.989±0.010** | **0.991±0.004** |
| | CSAL | 0.918±0.024 | 0.955±0.010 | 0.972±0.009 | 0.983±0.007 | 0.988±0.003 |
| | SMCDMAL | 0.929±0.021 | 0.955±0.013 | **0.977±0.009** | 0.982±0.007 | 0.985±0.006 |
| | PMCQAL | 0.880±0.050 | 0.925±0.050 | 0.964±0.014 | 0.983±0.008 | **0.988±0.005** |
| | MCDMAL | **0.954±0.012** | **0.973±0.007** | **0.981±0.007** | 0.986±0.005 | **0.988±0.004** |
| | **RMQCAL** | **0.958±0.007** | **0.973±0.016** | **0.983±0.013** | **0.990±0.003** | **0.991±0.001** |
| **LetterMN** | RANDOM | 0.977±0.010 | **0.992±0.002** | 0.994±0.003 | 0.996±0.002 | 0.997±0.001 |
| | MARGIN | 0.964±0.040 | **0.991±0.014** | **0.999±0.000** | 0.999±0.000 | 0.999±0.000 |
| | CLUSTER | 0.971±0.017 | **0.986±0.009** | 0.994±0.003 | 0.997±0.002 | 0.998±0.001 |
| | IDE | 0.969±0.017 | **0.988±0.007** | 0.997±0.002 | 0.998±0.001 | 0.998±0.001 |
| | DUAL | 0.950±0.025 | 0.972±0.011 | 0.974±0.007 | 0.980±0.008 | 0.983±0.007 |
| | QUIRE | **0.986±0.007** | **0.996±0.003** | 0.998±0.001 | 0.999±0.000 | 0.999±0.000 |
| | CSAL | 0.929±0.044 | 0.976±0.007 | 0.989±0.005 | 0.993±0.006 | 0.996±0.003 |
| | SMCDMAL | 0.942±0.027 | **0.989±0.006** | 0.998±0.003 | **1.000±0.001** | 1.000±0.001 |
| | PMCQAL | 0.872±0.035 | 0.953±0.024 | 0.987±0.009 | 0.998±0.001 | 0.999±0.000 |
| | MCDMAL | 0.929±0.024 | 0.959±0.022 | 0.969±0.016 | 0.983±0.009 | 0.994±0.002 |
| | **RMQCAL** | **0.973±0.010** | **0.990±0.008** | 0.997±0.001 | 0.998±0.001 | 0.998±0.000 |
| **LetterUV** | RANDOM | 0.992±0.005 | 0.996±0.004 | 0.998±0.001 | 0.999±0.000 | **1.000±0.000** |
| | MARGIN | 0.998±0.002 | **1.000±0.000** | **1.000±0.000** | **1.000±0.000** | **1.000±0.000** |
| | CLUSTER | 0.990±0.008 | 0.996±0.009 | **1.000±0.000** | **1.000±0.000** | **1.000±0.000** |
| | IDE | 0.995±0.004 | 0.999±0.001 | **1.000±0.000** | **1.000±0.000** | **1.000±0.000** |
| | DUAL | 0.983±0.014 | 0.986±0.008 | 0.990±0.008 | 0.991±0.008 | 0.993±0.007 |
| | QUIRE | **0.999±0.001** | **1.000±0.000** | **1.000±0.000** | **1.000±0.000** | **1.000±0.000** |
| | CSAL | 0.992±0.005 | 0.997±0.002 | **1.000±0.000** | 1.000±0.000 | **1.000±0.000** |
| | SMCDMAL | 0.992±0.006 | 0.998±0.002 | **1.000±0.000** | 1.000±0.000 | **1.000±0.000** |
| | PMCQAL | 0.987±0.008 | 0.997±0.003 | **1.000±0.000** | 1.000±0.000 | **1.000±0.000** |
| | MCDMAL | 0.992±0.004 | 0.998±0.001 | **1.000±0.001** | 1.000±0.001 | **1.000±0.001** |
| | **RMQCAL** | **0.994±0.004** | 0.999±0.001 | **1.000±0.001** | **1.000±0.001** | **1.000±0.001** |

**Table 8**

Win/Tie/Loss Counts of RMQCAL and Other Methods using Paired t-Tests for the 10 datasets

| | Percentage of the unlabeled samples | | | | | |
|---|---|---|---|---|---|---|
| | 5% | 10% | 20% | 25% | 30% | IN ALL |
| Algorithms | WIN/TIE/LOSS | WIN/TIE/LOSS | WIN/TIE/LOSS | WIN/TIE/LOSS | WIN/TIE/LOSS | WIN/TIE/LOSS |
| RANDOM | 4/6/0 | 5/4/1 | 8/1/1 | 8/1/1 | 7/2/1 | 32/14/4 |
| MARGIN | 5/4/1 | 4/4/2 | 3/4/3 | 2/5/3 | 2/5/3 | 16/22/12 |
| CLUSTER | 4/6/0 | 4/6/0 | 5/4/1 | 5/4/1 | 3/5/2 | 21/25/4 |
| IDE | 4/6/0 | 2/6/2 | 5/4/1 | 5/4/1 | 5/4/1 | 21/24/5 |
| DUAL | 8/2/0 | 8/1/1 | 9/1/0 | 8/1/1 | 8/1/1 | 41/6/3 |
| QUIRE | 3/5/2 | 2/5/3 | 2/6/2 | 2/5/3 | 3/4/3 | 12/25/13 |
| CSAL | 7/3/0 | 9/1/0 | 7/3/0 | 7/3/0 | 6/3/1 | 36/13/1 |
| SMCDMAL | 6/4/0 | 4/6/0 | 3/7/0 | 5/4/1 | 5/4/1 | 23/25/2 |
| PMCQAL | 7/2/1 | 6/4/0 | 7/3/0 | 5/4/1 | 4/4/2 | 29/17/4 |
| MCDMAL | 4/5/1 | 4/6/0 | 5/5/0 | 7/3/0 | 6/4/0 | 26/23/1 |
| *IN ALL* | 52/43/5 | 48/43/9 | 54/38/8 | 54/34/12 | 49/36/15 | 257/194/49 |

*3.5.5. Comparison between RMQCAL and state-of-the-art AL methods in large-scale datasets*

**Experiment F** is carried out to validate the effectiveness of the proposed method on a large-scale dataset. However, most contrast methods performed in **Experiment E** are too time consuming and do not work on large-scale datasets under insufficient hardware configurations (as shown in Table 11). In consideration of our limited experimental conditions, we made the following modification: the SQCs involved in our RMQCAL method are still DIVERSITY and MARGIN in RBF-SVM and QBC, but the contrast methods only include DIVERSITY and MARGIN in RBF-SVM, QBC, SMQCAL, CSAL and MCDMAL. The experiments corresponding to each method are repeated 10 times, and performance curves of the involved methods, which accuracy (X-axis) against labeling cost (Y-axis), are displayed in Fig. 21. Furthermore, two additional detailed AUC comparisons are made between RMQCAL and other MQCALs using these large-scale datasets according to a scheme similar to that shown in **Experiment E**; the experimental results of these comparisons are presented in Table 9 and Table 10.

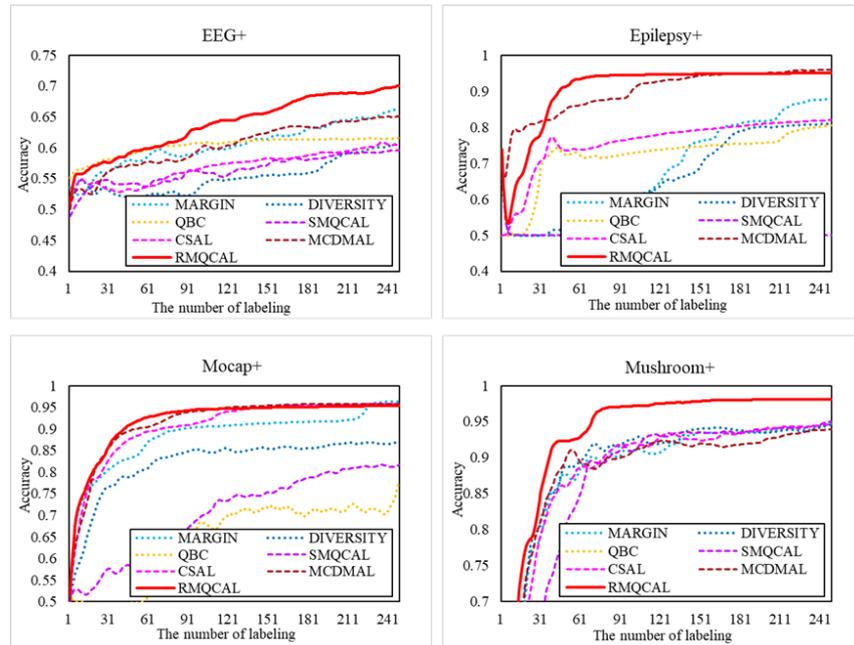

**Fig. 21.** Comparison of the accuracy of the 4 large-scale datasets

Table 9

Comparison of the AUC values of the 4 large-scale datasets

| | | Number of the unlabeled samples | | | | |
|---|---|---|---|---|---|---|
| | | 5 | 65 | 125 | 185 | 245 |
| Database | Algorithms | Mean±SD | Mean±SD | Mean±SD | Mean±SD | Mean±SD |
| Mushroom+ | MARGIN | 0.492±0.032 | **0.625±0.042** | **0.657±0.052** | **0.697±0.056** | **0.726±0.072** |
| | DIVERSITY | **0.533±0.063** | **0.566±0.036** | **0.608±0.029** | 0.616±0.032 | 0.636±0.044 |
| | QBC | **0.511±0.050** | 0.531±0.050 | 0.542±0.041 | 0.545±0.041 | 0.551±0.045 |
| | SMCDMAL | **0.525±0.050** | **0.585±0.038** | 0.604±0.036 | 0.610±0.027 | 0.618±0.032 |
| | CSAL | **0.512±0.049** | **0.586±0.049** | **0.615±0.062** | 0.627±0.080 | 0.650±0.091 |
| | MCDMAL | **0.539±0.043** | **0.624±0.043** | **0.655±0.049** | **0.679±0.055** | **0.705±0.062** |
| | **RMQCAL** | **0.537±0.042** | **0.607±0.061** | **0.659±0.072** | **0.713±0.087** | **0.752±0.069** |
| EEG+ | MARGIN | **0.985±0.006** | 0.985±0.005 | 0.985±0.007 | 0.986±0.007 | 0.989±0.005 |
| | DIVERSITY | **0.984±0.006** | 0.984±0.006 | 0.984±0.006 | 0.986±0.006 | 0.988±0.006 |
| | QBC | 0.866±0.031 | 0.964±0.035 | **0.969±0.037** | **0.969±0.037** | 0.975±0.023 |
| | SMCDMAL | **0.984±0.007** | 0.984±0.007 | 0.984±0.007 | 0.984±0.007 | 0.984±0.007 |
| | CSAL | 0.866±0.047 | 0.978±0.005 | 0.984±0.004 | 0.984±0.005 | 0.986±0.003 |
| | MCDMAL | **0.984±0.008** | 0.985±0.004 | 0.989±0.004 | **0.992±0.002** | 0.994±0.002 |
| | **RMQCAL** | **0.986±0.007** | 0.990±0.002 | 0.992±0.001 | 0.994±0.002 | 0.994±0.002 |
| Mocap+ | MARGIN | 0.855±0.086 | 0.919±0.093 | 0.939±0.081 | 0.943±0.079 | 0.973±0.014 |
| | DIVERSITY | 0.803±0.098 | 0.898±0.008 | 0.914±0.008 | 0.922±0.010 | 0.929±0.007 |
| | QBC | 0.699±0.133 | 0.918±0.021 | 0.937±0.009 | 0.945±0.012 | 0.949±0.011 |
| | SMCDMAL | 0.756±0.116 | 0.840±0.072 | 0.915±0.030 | 0.934±0.008 | 0.943±0.004 |
| | CSAL | 0.856±0.045 | **0.939±0.029** | **0.961±0.007** | **0.967±0.010** | **0.969±0.010** |
| | MCDMAL | **0.818±0.150** | **0.938±0.044** | **0.966±0.011** | **0.969±0.013** | **0.970±0.012** |
| | **RMQCAL** | **0.900±0.017** | **0.955±0.009** | **0.963±0.012** | **0.966±0.012** | **0.969±0.012** |
| Epilepsy+ | MARGIN | **0.889±0.037** | **0.992±0.003** | **0.996±0.002** | **0.997±0.002** | **0.997±0.002** |
| | DIVERSITY | **0.858±0.082** | **0.991±0.004** | **0.995±0.002** | **0.997±0.002** | **0.997±0.002** |
| | QBC | **0.891±0.076** | 0.977±0.020 | 0.986±0.010 | 0.992±0.005 | 0.994±0.003 |
| | SMCDMAL | 0.693±0.124 | **0.993±0.004** | **0.998±0.002** | **0.998±0.001** | **0.999±0.001** |
| | CSAL | **0.755±0.172** | **0.991±0.009** | **0.994±0.007** | **0.998±0.002** | **0.998±0.001** |
| | MCDMAL | **0.860±0.062** | **0.992±0.003** | **0.995±0.003** | **0.997±0.001** | **0.998±0.001** |
| | **RMQCAL** | **0.820±0.116** | **0.988±0.009** | **0.996±0.005** | **0.996±0.005** | **0.996±0.004** |

Table 10

Win/Tie/Loss Counts of RMQCAL versus Other Methods using Paired t-Tests for the 4 large-scale datasets

| | Number of the unlabeled samples | | | | | |
|---|---|---|---|---|---|---|
| | 5 | 65 | 125 | 185 | 245 | IN ALL |
| Algorithms | WIN/TIE/LOSS | WIN/TIE/LOSS | WIN/TIE/LOSS | WIN/TIE/LOSS | WIN/TIE/LOSS | WIN/TIE/LOSS |
| MARGIN | 1/3/0 | 1/3/0 | 1/3/0 | 1/3/0 | 1/3/0 | 5/15/0 |
| DIVERSITY | 1/3/0 | 2/2/0 | 2/2/0 | 3/1/0 | 3/1/0 | 11/9/0 |
| QBC | 2/2/0 | 3/1/0 | 3/1/0 | 2/2/0 | 3/1/0 | 13/7/0 |
| SMCDMAL | 2/2/0 | 2/2/0 | 3/1/0 | 3/1/0 | 3/1/0 | 13/7/0 |
| CSAL | 2/2/0 | 1/3/0 | 1/3/0 | 2/2/0 | 2/2/0 | 8/12/0 |
| MCDMAL | 0/4/0 | 1/3/0 | 1/3/0 | 0/4/0 | 0/4/0 | 2/18/0 |
| *IN ALL* | 8/16/0 | 10/14/0 | 11/13/0 | 11/13/0 | 12/12/0 | 52/68/0 |

From **Experiment F**, it can also be observed that the proposed RMQCAL also has the best performance among all feasible methods in a large-scale database. Although the advantage of the proposed RMOCAL on AUC is not very large according to a t-test, the proposed method still has the ability to yield a highly stable and accurate solution with reduced labeling cost. Except for EEG, the performance curve grows and flattens with less than 100 samples, which also demonstrates the potential value of AL algorithms in big-data problems.

*3.5.6. Comparing the CPU time for our RMQCAL and another AL method*

In **Experiment G**, each involved method is run ten times; the average CPU time for each query in each method is recorded (in seconds), as shown in Table 11. The involved methods can be divided into three types: '*single-query criterion-based AL*', '*state-of-the-art MQCAL*,' and '*the proposed RMQCAL*'. There are two points worth mentioning. First, 'Error' means that the corresponding AL methods are inestimable and cannot be used in this dataset under our experimental conditions. Second, the SQCs involved in our RMQCAL method are Diversity and Margin in RBF-SVM and QBC.

Table 11

Comparison of CPU time between our RMQCAL method and another AL method

| | Single query criterion AL | | | | state-of-the-art MQCAL | | | | the proposed RMQCAL | | |
|---|---|---|---|---|---|---|---|---|---|---|---|
| Database | Diversity | QBC | Margin | CSAL | SMCQ-AL | PMCQ-AL | MCDM-AL | QUIRE | BORD-A | BUCK-LIN | MARK-OV |
| Austra | 0.5000 | 0.0938 | 0.0155 | 0.6406 | 0.0469 | 1.5000 | 1.0313 | 0.3125 | **0.6406** | **0.6875** | **0.7188** |
| Isolet | 0.5165 | 0.4688 | 0.0690 | 1.0313 | 0.0625 | 2.0938 | 1.1563 | 0.2500 | **1.0313** | **1.1250** | **1.1094** |
| Titato | 0.7500 | 0.1094 | 0.0313 | 1.1094 | 0.0469 | 2.9375 | 1.6094 | 1.4532 | **0.8750** | **0.9219** | **0.9063** |
| Vehicle | 0.3125 | 0.0938 | 0.0031 | 0.4688 | 0.0156 | 0.5000 | 0.5313 | 0.1094 | **0.3906** | **0.3750** | **0.4063** |
| Wdbc | 0.4219 | 0.0781 | 0.0063 | 0.6875 | 0.0156 | 1.2500 | 0.8125 | 0.1719 | **0.5156** | **0.5313** | **0.5469** |
| LetterDP | 1.4063 | 0.1563 | 0.0313 | 1.8281 | 0.0313 | 9.7656 | 3.7031 | 7.8750 | **1.4688** | **1.5000** | **1.4844** |
| LetterEF | 1.2656 | 0.1563 | 0.0343 | 1.5000 | 0.0625 | 8.8906 | 3.2301 | 6.9500 | **1.4063** | **1.4688** | **1.4219** |
| LetterIJ | 1.1875 | 0.1250 | 0.0469 | 1.7500 | 0.0313 | 9.8705 | 3.0625 | 6.3500 | **1.3594** | **1.3750** | **1.4063** |
| LetterMN | 1.2188 | 0.1719 | 0.0328 | 1.8750 | 0.0469 | 10.1719 | 3.4375 | 7.4688 | **0.9111** | **0.9187** | **0.9288** |
| LetterUV | 1.2813 | 0.2031 | 0.0313 | 1.5781 | 0.0156 | 10.9531 | 3.5938 | 7.4688 | **0.9505** | **0.9657** | **0.9708** |
| EEG | 14.7656 | 1.3281 | 0.4688 | 16.1094 | 0.3906 | Error | 37.0625 | Error | **13.6700** | **13.6250** | **13.7980** |
| Epileptic | 3.9688 | 0.8438 | 0.0938 | 5.5938 | 0.1250 | 150.650 | 7.0938 | 242.540 | **4.4844** | **5.1250** | **5.0496** |
| MoCap | 26.6719 | 1.9531 | 1.5469 | 35.7188 | 1.7656 | Error | 185.578 | Error | **30.1563** | **28.5781** | **28.9375** |
| Mushroom | 6.3281 | 0.6250 | 0.1719 | 13.5000 | 0.1875 | Error | 13.5000 | Error | **7.2500** | **7.2656** | **7.2344** |

The following conclusions are obtained from the results of **Experiment G**. (1) Unsurprisingly, all single criterion-based AL methods are more efficient than RMQCAL because they are also components of our RMQCAL, and the CPU time of RMQCAL is approximately equal to the sum of the times spent by each constituent SQC. (2) As expected, the Markov chain does not require more CPU time than Borda's and Bucklin's methods with the added step of 'sample truncation'. (3) Compared with MQCAL, RMQCAL is comparatively efficient and only second to SMQCAL. We believe that this result is acceptable because the efficiency of SMQCAL comes at the cost of performance. The multi-layer filter, like the design of SMQCAL, can, indeed, significantly reduce the operational time. However, such a design will miss many of the samples with high comprehensive values, leadings to a suboptimal result, as shown in **Experiments E** and **F**. (4) For the large-scale database, this article does not recommend the use of QUIRE and PMQCAL. Both of these are too inefficient and may even fail to work when the operational environment is not adequately established.

**4. Conclusion**

In this paper, a means is presented for training data selection in AL problems. Unlike conventional

AL methods, it can be ensured that the samples selected for labeling are overall valuable because multiple SQC are involved in the proposed method, and they are combined by the introduction of a weighted rank aggregation.

The proposed RMQCAL avoids building a multilayered filter-like process or solving complex optimization equations, and this capability is highlighted as the main contribution of this study. With respect to advantages, the proposed RMQCAL favorably inherits the merits of most existing MQCAL methods. When applying our RMQCAL, less human intervention is required and fewer empirical parameters are used, and any number and type of SQC can be used and blended into one through dynamical weighting.

To achieve optimal performance, several combinations of SQC adapted from conventional AL methods were applied. Moreover, existing ranking aggregation methods (e.g., Borda's, Bucklin voting and Markov chain methods) were improved as a key facet of our RMQCAL process. In addition to applying these methods, we employed other ranking aggregation methods, including Thurstone's model, the cross-entropy Monte Carlo model, and the Condorcet model. However, as some methods oppose the AL method in theory or have a run time for realizing one AL iteration that is too long, these methods are not used in our MQCAL. Nevertheless, other more appropriate methods may exist.

Our experimental results show that our newly designed RMQCAL is more effective than the conventional SQC-based AL method. Relative to other MQCAL models, RMQCAL is also rated among the best. Either for a conventional classification task or a large-scale data classification task, the proposed RMQCAL has the ability to do well in helping users train a superior classification model with fewer labeling costs and less running time. Moreover, RMQCAL, in our view, can be an appropriate solution for practical issues, especially when there is no validation set in hand and the labeling cost of each sample is very expensive.

Our planned future work will focus on three main points. First, we will attempt to extend our method to more complex classification or regression problems, e.g., multiclass and multi-labeled problems, and our latest research indicates that RMQCAL also performs well in ordinal regression. Second, the theoretical proof of RMQCAL should be studied further. Finally, we will attempt to apply this approach to medical lesion recognition.

**Acknowledgments**

This research is partially supported by the National Key Research and Development Program (2016YFC0106200), the 863 National Research Fund (2015AA043203), and the Chinese NSFC Research Fund (61190120, 61190124 and 61271318). Special funding was received from capital health research and development under grant No. 2016-1-4011. The authors are grateful to the University of South Florida and Sandia National Laboratories, who provide UCI as a resource for our experimental data. We also express our sincere gratitude to the Department of Computer Science & Technology of Nanjing University for providing their experimental results for comparison in this paper.

**References**

[1]   B. Settles, Active learning literature survey, University of Wisconsin, Madison. 15 (2010) 201–221. doi:10.1.1.167.4245.

[2]   I. Muslea, S. Minton, C.A. Knoblock, Active learning with multiple views, Journal of Artificial Intelligence Research. 27 (2006) 203–233. doi:10.1613/jair.2005.

[3]   N. Panda, K.-S. Goh, E.Y. Chang, Active learning in very large databases, Multimedia Tools

and Applications. 31 (2006) 249–267. doi:10.1007/s11042-006-0043-1.

[4] Z. Xu, K. Yu, V. Tresp, X. Xu, J. Wang, Representative Sampling for Text Classification Using Support Vector Machines, in: F. Sebastiani (Ed.), Advances in Information Retrieval. Springer Berlin Heidelberg, Berlin, Heidelberg, 2003: pp. 393–407.

[5] H. Huang, C. Zhang, Q. Hu, P. Zhu, Multi-view representative and informative induced active learning, in: Lecture Notes in Computer Science (including subseries Lecture Notes in Artificial Intelligence and Lecture Notes in Bioinformatics), 2016: pp. 139–151. doi:10.1007/978-3-319-42911-3_12.

[6] E. Lughofer, Hybrid active learning for reducing the annotation effort of operators in classification systems, Pattern Recognition. 45 (2012) 884–896. doi: 10.1016/j.patcog.2011.08.009.

[7] P. Donmez, J.G. Carbonell, P.N. Bennett, Dual Strategy Active Learning, Machine Learning ECML 2007. (2007) 116–127. doi:10.1007/978-3-540-74958-5_14.

[8] S.J. Huang, R. Jin, Z.H. Zhou, Active Learning by Querying Informative and Representative Examples, IEEE Transactions on Pattern Analysis and Machine Intelligence. 36 (2014) 1936–1949. doi:10.1109/TPAMI.2014.2307881.

[9] R. Wang, S. Kwong, Active learning with multi-criteria decision making systems, Pattern Recognition. 47 (2014) 3106–3119. doi:10.1016/j.patcog.2014.03.011.

[10] S. Dasgupta, D. Hsu, Hierarchical sampling for active learning, in: Proceedings of the 25th international conference on Machine learning - ICML '08, 2008: pp. 208–215. doi:10.1145/1390156.1390183.

[11] Y. Jiao, P. Zhao, J. Wu, Y. Shi, Z. Cui, A Multicriterion Query-Based Batch Mode Active Learning Technique, in: Foundations of Intelligent Systems, Springer, 2014: pp. 669–680.

[12] S. Jiang, O.U. Qing-Yu, Batch-mode active learning approach of computer viruses classifier based on information density, J. Nav. Univ. Eng. (2015).

[13] K. Yu, J. Bi, V. Tresp, Active learning via transductive experimental design, in: Proceedings of the 23rd international conference on Machine learning ICML 06. 148 (2006) 1081–1088. doi:10.1145/1143844.1143980.

[14] D. Cai, X. He, Manifold adaptive experimental design for text categorization, IEEE Transactions on Knowledge and Data Engineering. 24 (2012) 707–719. doi:10.1109/TKDE.2011.104.

[15] C.C. Chang, B.H. Liao, Active learning based on minimization of the expected path-length of random walks on the learned manifold structure, Pattern Recognition. 71 (2017) 337–348. doi: 10.1016/j.patcog.2017.06.001.

[16] Z. Wang, S. Yan, C. Zhang, Active learning with adaptive regularization, Pattern Recognition. 44 (2011) 2375–2383. doi:10.1016/j.patcog.2011.03.008.

[17] A. Holub, P. Perona, M.C. Burl, Entropy-based active learning for object recognition, in: 2008 IEEE Computer Society Conference on Computer Vision and Pattern Recognition Workshops, 2008: pp. 1–8. doi:10.1109/CVPRW.2008.4563068.

[18] N. Roy, A. Mccallum, M.W. Com, Toward optimal active learning through monte carlo estimation of error reduction., Proceedings of the International Conference on Machine Learning (ICML). (2001) 441–448.

[19] Y. Freund, H.S. Seung, E. Shamir, N. Tishby, Selective Sampling Using the Query by Committee Algorithm, Machine Learning. 168 (1997) 133–168. doi:10.1.1.20.8521.

[20] Q. Zhang, S. Sun, Multiple-view multiple-learner active learning, Pattern Recognition. 43 (2010) 3113–3119. doi: 10.1016/j.patcog.2010.04.004.

[21] Y. Baram, R. El-Yaniv, K. Luz, Online Choice of Active Learning Algorithms, Journal of Machine Learning Research. 5 (2004) 255–291. doi:10.1017/CBO9781107415324.004.

[22] P. Auer, N. Cesa-Bianchi, Y. Freund, R.E. Schapire, Gambling in a rigged casino: The adversarial multi-armed bandit problem, in: Foundations of Computer Science, 1995. Proceedings., 36th Annual Symposium on, 1995: pp. 322–331.

[23] D. Shen, J. Zhang, J. Su, G. Zhou, C.-L. Tan, Multi-Criteria-based Active Learning for Named Entity Recognition, in: Proceedings of the 42nd Meeting of the Association for Computational Linguistics (ACL'04), (2004) 589–596. doi:10.3115/1218955.1219030.

[24] S. Patra, K. Bhardwaj, L. Bruzzone, A Spectral-Spatial Multicriteria Active Learning Technique for Hyperspectral Image Classification, IEEE Journal of Selected Topics in Applied Earth Observations and Remote Sensing, 10 (2017) 5213-5227. doi: 10.1109/JSTARS.2017.2747600

[25] B. Demir, C. Persello, L. Bruzzone, Batch-mode active-learning methods for the interactive classification of remote sensing images, IEEE Transactions on Geoscience and Remote Sensing. 49 (2011) 1014–1031. doi:10.1109/TGRS.2010.2072929.

[26] S. Patra, L. Bruzzone, A novel SOM-SVM-based active learning technique for remote sensing image classification, IEEE T Geosci Remote, 52 (2014) 6899-6910.
doi: 10.1109/TGRS.2014.2305516

[27] B. Demir, L. Bruzzone, A multiple criteria active learning method for support vector regression, Pattern recognition, 47 (2014) 2558-2567. doi: 10.1016/j.patcog.2014.02.001

[28] C.T. Symons, N.F. Samatova, R. Krishnamurthy, B.H. Park, T. Umar, D. Buttler, T. Critchlow, D. Hysom, Multi-criterion active learning in conditional random fields, Tools with Artificial Intelligence, 2006. ICTAI'06. 18th IEEE International Conference on, IEEE2006, pp. 323-331
doi: 10.1109/ICTAI.2006.90

[29] B. Du, Z. Wang, L. Zhang, L. Zhang, W. Liu, J. Shen, D. Tao, Exploring representativeness and informativeness for active learning, IEEE transactions on cybernetics, 47 (2017) 14-26.
doi: 10.1109/TCYB.2015.2496974

[30] A. Samat, P. Gamba, S. Liu, P. Du, J. Abuduwaili, Jointly informative and manifold structure representative sampling based active learning for remote sensing image classification, IEEE T Geosci Remote, 54 (2016) 6803-6817
DOI: 10.1109/TGRS.2016.2591066

[31] X. Xu, J. Li, S. Li, Multiview Intensity-Based Active Learning for Hyperspectral Image Classification, IEEE T Geosci Remote, 56 (2018) 669-680.
DOI: 10.1109/TGRS.2017.2752738

[32] J. Zhang, D. Chen, H. Xie, S. Zhang, L. Gu, I don't know: Double-strategies based active learning for mammographic mass classification, Life Sciences Conference (LSC), 2017 IEEE, IEEE2017, pp. 182-185.
DOI: 10.1109/LSC.2017.8268173

[33] S. Lin, Rank aggregation methods, Wiley Interdisciplinary Reviews: Computational Statistics. 2 (2010) 555–570. doi:10.1002/wics.111.

[34] L. Zheng, S. Wang, L. Tian, F. He, Z. Liu, Q. Tian, Query-adaptive late fusion for image search and person re-identification, in: Proceedings of the IEEE Computer Society Conference on Computer Vision and Pattern Recognition, 2015: pp. 1741–1750.

doi:10.1109/CVPR.2015.7298783.

[35] T.A. Solgård, P. Landskroener, Municipal voting system reform: overcoming the legal obstacles, Choice. 1 (2002) 3rd.

[36] C. Dwork, R. Kumar, M. Naor, D. Sivakumar, Rank aggregation methods for the web, in: Proceedings of the 10th international conference on World Wide Web, 2001: pp. 613–622.

[37] H.P. Young, Condorcet's Theory of Voting, American Political Science Review. 82 (1988) 1231–1244. doi:10.2307/1961757.

[38] S. Niu, Y. Lan, J. Guo, X. Cheng, Stochastic rank aggregation, arXiv preprint arXiv:1309.6852. (2013).

[39] M.G. Kendall, Rank correlation methods., 1948.

[40] P. Diaconis, R.L. Graham, Spearman's Footrule as a Measure of Disarray, Journal of the Royal Statistical Society. Series B (Methodological). 39 (1977) 262--268. doi:10.2307/2984804.

[41] S. Tong, D. Koller, Support Vector Machine Active Learning with Applications to Text Classification, Journal of Machine Learning Research. (2001) 45–66. doi:10.1162/153244302760185243.

[42] S.C.H. Hoi, M.R. Lyu, Semi-supervised SVM batch mode active learning for image retrieval, 2008 IEEE Conference on Computer Vision and Pattern Recognition. (2008) 1–7. doi:10.1109/CVPR.2008.4587350.